\documentclass{article} 
\usepackage{preprint,times}

\usepackage{amsmath,amsfonts,bm}

\def\eqref#1{equation~\ref{#1}}

\def\1{\bm{1}}

\DeclareMathAlphabet{\mathsfit}{\encodingdefault}{\sfdefault}{m}{sl}
\SetMathAlphabet{\mathsfit}{bold}{\encodingdefault}{\sfdefault}{bx}{n}

\usepackage{hyperref}
\usepackage{url}

\usepackage{times}  
\usepackage{helvet}  
\usepackage{courier}  

\usepackage[ruled,vlined]{algorithm2e}

\usepackage{newfloat}
\usepackage{listings}
\usepackage{bibentry}

\usepackage{xcolor}
\definecolor{lightgray}{gray}{0.92}

\usepackage[most]{tcolorbox}
\usepackage{fvextra}
\usepackage{cuted}

\usepackage{listings}
\usepackage[ruled,vlined]{algorithm2e}
\usepackage{caption}
\usepackage{float}
\usepackage{arydshln}

\usepackage{booktabs}
\usepackage{array}

\title{InfoAgent: Traceable Generation and Repair of Evidence-Grounded Infographics}

\author{
\textbf{
Yifan Li\textsuperscript{1,2},
Tong Li\textsuperscript{2},
Qi Zeng\textsuperscript{2},
Lishuai Gao\textsuperscript{2,$\ddagger$},
Ruwei Pan\textsuperscript{3},
Cong Wei\textsuperscript{2},
}\\
\textbf{
Shaohua Kevin Zhou\textsuperscript{1}
Zhuoliang Kang\textsuperscript{2},
Xiaoming Wei\textsuperscript{2},
} \\
\textsuperscript{1}University of Science and Technology of China \quad
\textsuperscript{2}Meituan \quad
\textsuperscript{3}Peking University, Peking University \\
\textsuperscript{$\ddagger$}Project lead \\
\texttt{liyifancqu@163.com}
}

\arxivcopy 
\renewcommand{\headrulewidth}{0pt}
\renewcommand{\footrulewidth}{0pt}
\AddToShipoutPicture{%
  \AtPageLowerLeft{%
    \raisebox{0.34in}{\makebox[\paperwidth][c]{\thepage}}%
  }%
}

\usepackage{graphicx}
\usepackage{titletoc}
\usepackage{amsmath}
\usepackage{amssymb}
\usepackage{multirow}
\usepackage{xcolor}
\usepackage{xspace}
\usepackage{cleveref}
\usepackage{booktabs}

\crefname{figure}{Figure}{Figures}
\crefname{table}{Table}{Tables}
\crefname{section}{Sec.}{Secs.}
\Crefname{section}{Section}{Sections}
\crefname{equation}{Eq.}{Eqs.}

\usepackage{natbib}  
\usepackage{caption} 
\begin{document}

\maketitle

\begin{abstract}
Reliable infographic generation requires facts, symbols, and visual relations to remain consistent through rendering and revision. Correcting one element also requires tracking its supporting evidence and the dependencies affected by the change. We present \textbf{InfoAgent}, a training-free framework for \emph{evidence-bound visual-symbolic program synthesis}. Its Infographic Visual Description (IVD) records factual payloads, evidence provenance, execution routes, and verification obligations in a typed dependency graph. Retrieved design priors guide compilation, and layered execution combines raster synthesis with editable symbolic and binding objects while retaining their traces. Dependency-aware repair localizes corrections, rechecks affected dependencies, and requires protected obligations to remain satisfied under the declared checkers. Unresolved obligations remain explicit. On IGenBench, InfoAgent achieves 93.0 Q-ACC and 59.0 I-ACC. We also introduce InfoGraphicBench-Evidence, where complete-checklist pass rates on 200 test requests increase from 21.5\% for Same-IVD Prompt to 23.5\% for the initial layered output and 28.5\% after repair, using the same evidence and initial IVD. On 120 audited repair cases, localized repair edits 12.4\% of the canvas on average, compared with 67.3\% for global regeneration.
\end{abstract}
\section{Introduction}
\label{sec:introduction}

Infographics combine factual claims, numerical data, and visual explanations on one canvas. In journalism, education, and public communication, their usefulness depends on readable design and accurate information~\cite{feng2026infoalign,cui2019text,tyagi2022infographics}. A polished composition can still mislead when a value is unsupported, a source note is missing, or a label refers to the wrong object~\cite{wang2019datashot,vu2025factflow,tang2026igenbench}. Reliable generation requires preserving relationships among evidence, text, and visual elements, not just their appearance.

Infographic research now spans document-conditioned generation, authoring tools, chart resources, and reliability evaluation~\cite{li2025chartgalaxy,ghosh2025infogen,tang2026igenbench}. Image models offer increasingly capable synthesis and text rendering~\cite{esser2024scaling,flux2024,betker2023improving,wu2025qwen,cai2025z,cai2025hidream,team2025longcat}. Research agents acquire and organize external knowledge~\cite{jin2025search,zheng2025deepresearcher,yang2026multimodal,chng2025sensenova}, and image-generation agents use search, references, planning, and feedback to improve contextual and compositional fidelity~\cite{zhang2026qwen,feng2026gen,chen2026genevolve,chen2026unify,li2026codrawagents}. Structured execution and local correction are also established approaches: InfoGen converts document-derived metadata into infographic code~\cite{ghosh2025infogen}, GenClaw uses executable visual sketches before raster synthesis~\cite{ye2026genclaw}, and M3 checks constraints and validates targeted image edits~\cite{yang2026m3}.

We study how to preserve evidence and execution dependencies as an infographic is revised. Consider a percentage callout whose text is correct but whose arrow points to the wrong chart segment. Correcting the arrow requires locating its intended target; moving the callout may obscure a neighboring value or source note. An editable object provides a location for the change, but does not by itself specify the supporting evidence or the dependent requirements that need rechecking. A reliable repair needs these relationships to remain explicit across planning, rendering, and verification.

We introduce \textbf{InfoAgent}, a training-free framework for \emph{evidence-bound visual-symbolic program synthesis}. Its intermediate representation, \textbf{Infographic Visual Description} (IVD), records content, evidence spans, spatial regions, execution routes, bindings, and verification obligations in a typed dependency graph. The same element identifier connects a factual claim to its supporting span, rendered object, and post-composition checks. This evidence-linked form of \emph{element addressability} allows a detected failure to be traced to its content, geometry, or local relation. Retrieved layout and style priors guide the design, while factual content comes from supplied or retrieved evidence. All foundation models remain frozen.

As shown in \cref{fig:framework}, InfoAgent routes open-ended imagery to a visual generator, exact text and data graphics to symbolic rendering, and arrows, callouts, and legend associations to binding execution. Composition retains rendering-tree handles for symbolic objects and region-level or grounded associations for raster elements. Checkers use these traces to associate each \textsc{Pass}, \textsc{Fail}, or \textsc{Unknown} decision with an element and a diagnostic witness. Repair follows the dependency graph to identify affected elements and relevant checks. Candidate patches must preserve previously passed critical obligations and all passed obligations outside the target's dependency closure under the declared checkers. The resulting certificate records these checks and unresolved cases; semantic verification remains model-assisted.

We evaluate InfoAgent on the complete IGenBench and introduce \textbf{InfoGraphicBench-Evidence}, a 260-request benchmark with frozen evidence bundles and held-out annotations, including 200 test requests. InfoAgent achieves 93.0 Q-ACC and 59.0 I-ACC on IGenBench. On InfoGraphicBench-Evidence, the complete-checklist pass rate (Full) rises from 21.5\% for Same-IVD Prompt to 23.5\% for the initial layered output and 28.5\% after repair. These comparisons use the same evidence and initial IVD. On the audited repair subset, localized repair changes 12.4\% of the canvas on average, compared with 67.3\% for global regeneration. The results support the value of trace-preserving execution and scoped repair, while the remaining failures show that evidence quality and semantic verification continue to limit reliability.

We introduce an evidence-bound IVD that connects source spans, rendered elements, local relations, and verification obligations. Building on this representation, InfoAgent combines layered execution with dependency-aware repair to localize corrections, recheck affected content, and record unresolved requirements. We also construct InfoGraphicBench-Evidence and evaluate whole-infographic reliability, output quality, and repair locality alongside IGenBench.
\section{Related Work}
\label{sec:related_work}

\paragraph{Agentic image generation.}
Modern text-to-image models offer strong visual synthesis and increasingly capable instruction following and text rendering~\cite{esser2024scaling,flux2024,betker2023improving,wu2025qwen,cai2025z,cai2025hidream}. Recent agentic systems extend these models with external knowledge, tool use, and iterative control. Gen-Searcher retrieves textual evidence and visual references before generation~\cite{feng2026gen}, while Qwen-Image-Agent constructs missing generation context through planning, reasoning, search, memory, and feedback~\cite{zhang2026qwen}. M3 decomposes complex requests into checkable constraints and revises failed components~\cite{yang2026m3}; related approaches use grounded recaptioning, tool trajectories, specialized agents, or multi-round reasoning to improve factual and compositional generation~\cite{chen2026unify,chen2026genevolve,li2026codrawagents,bian2026rs}. GenClaw further introduces executable SVG, HTML, and code-based canvases before raster synthesis~\cite{ye2026genclaw}. InfoAgent complements these directions by binding each information-bearing element to evidence, an execution route, local relations, and checkable obligations. The resulting traces support element-indexed verification and scoped re-execution after composition, rather than treating the generated artifact only as a holistic image.

\paragraph{Structured visual communication.}
Deep-research and presentation systems organize evidence into multimodal reports or slides using visualization descriptions, reference layouts, editable code, and iterative refinement~\cite{zheng2025deepresearcher,yang2026multimodal,zheng2025pptagent,tang2025slidecoder,zheng2026deeppresenter,zeng2026slidetailor,xu2025pregenie}. Infographic research has progressed from constrained linguistic or tabular inputs to automatic data stories, visualization code, and AI-assisted authoring~\cite{cui2019text,wang2019datashot,shi2020calliope,tyagi2022infographics,dibia2023lida,vu2025factflow}. Recent work contributes large-scale infographic-chart resources~\cite{li2025chartgalaxy}, document-conditioned statistical infographic generation~\cite{ghosh2025infogen}, reliability evaluation~\cite{tang2026igenbench}, and narrative-centered co-creation~\cite{feng2026infoalign}. These systems provide complementary mechanisms for content selection, page organization, visual encoding, rendering, and assessment. InfoAgent focuses on open-ended, evidence-grounded single-canvas infographics, where factual claims, exact symbols, visual objects, and local bindings must remain jointly traceable. IVD connects these stages through an executable element graph, combining open-ended visual synthesis with editable symbolic and binding execution.
\section{Method}
\label{sec:method}

\begin{figure*}[t]
\centering
\includegraphics[width=\linewidth]{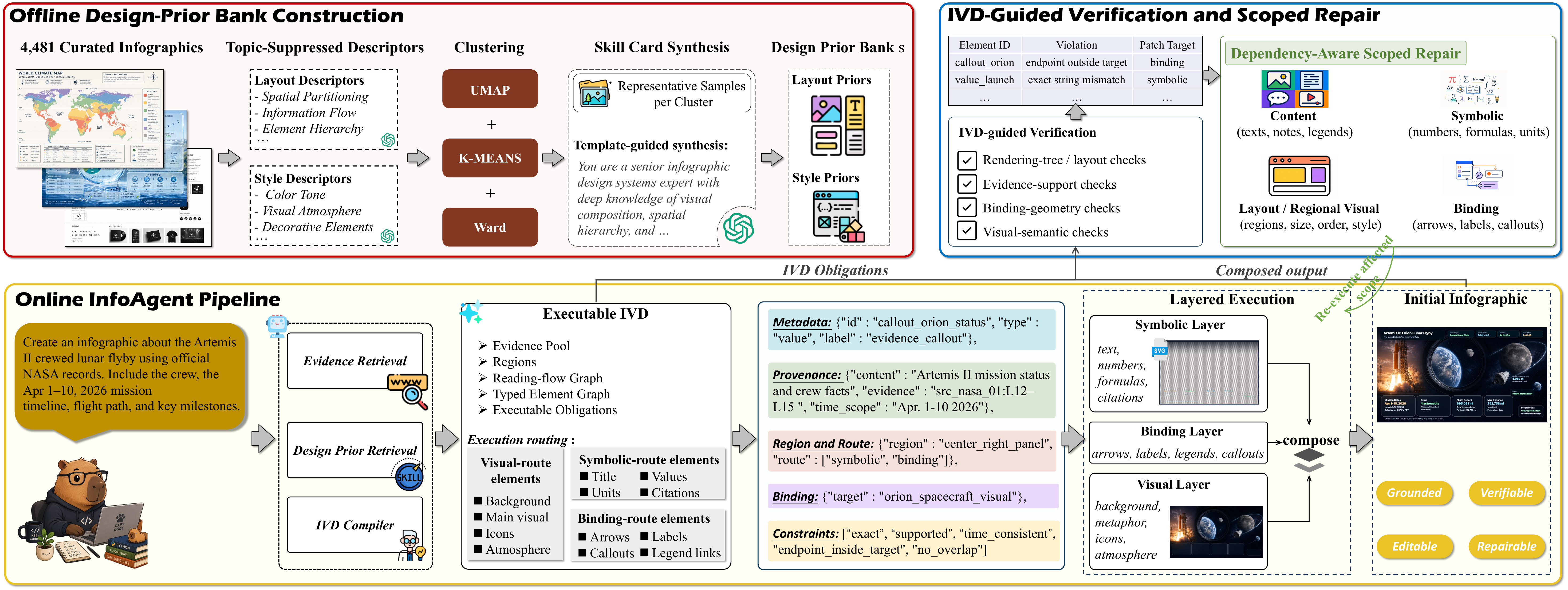}
\caption{
Overview of InfoAgent. A user request, retrieved evidence, and design
priors are compiled into a typed IVD. Each information-bearing element
retains evidence provenance, a spatial region, an execution route,
binding dependencies, and checkable obligations. Visual, symbolic, and
binding layers are composed with execution traces. IVD-guided
verification then maps element-indexed violations to scoped content,
symbolic, binding, layout, or visual patches.
}
\label{fig:framework}
\end{figure*}

We propose \textbf{InfoAgent}, a training-free framework that formulates knowledge-intensive infographic generation as \emph{evidence-bound visual-symbolic program synthesis}. Given a user request $x$, an evidence pool $\mathcal{E}$, and a retrieved design prior $s$, InfoAgent compiles them into an executable IVD and produces the composed infographic together with its execution trace:
\begin{equation}
z=F_{\mathrm{ivd}}(x,\mathcal{E},s),\qquad
(y,T)=\operatorname{Exec}(z),
\label{eq:overview}
\end{equation}
where $z$ is the compiled IVD, $y$ is the composed infographic, and $T$ records correspondences between IVD elements and rendered objects or regions. All foundation models and rendering tools remain frozen. \cref{fig:framework} summarizes the pipeline from IVD compilation to scoped repair.

Rather than serving as a longer prompt, IVD makes each information-bearing element addressable by recording its payload, evidence provenance, spatial role, execution route, binding and dependency relations, and post-composition obligations. InfoAgent executes these elements through visual, symbolic, and binding layers, preserving traces for checking and repair. This separation allows an incorrect value, source note, or binding relation to be re-executed without necessarily regenerating the complete artifact.

Verification is defined relative to the obligations declared by IVD. An output is marked as certified relative to $z$ only when every critical obligation returns \textsc{Pass}; a critical \textsc{Unknown} remains unresolved. The certificate records checker statuses, confidence values, and witnesses for the declared content, layout, and binding obligations. It provides an operational account of these declared requirements rather than a guarantee of open-world factual correctness.

\subsection{Evidence and Design Priors}
\label{sec:retrieval}

InfoAgent retrieves two complementary forms of context. The
\emph{evidence retriever} collects passages, numerical values, dates,
entity names, and source metadata. Retrieved documents are deduplicated
and segmented into addressable spans. Every non-decorative factual unit
must be linked to at least one span; otherwise, it is marked
\texttt{blocked} and is excluded from execution.

The \emph{design retriever} supplies layout and style constraints without
changing the factual payload. We construct a non-parametric prior bank
from 4,481 curated infographics. Topic-suppressed layout and style
descriptions are embedded and clustered using UMAP, K-Means, and Ward
merging~\cite{mcinnes2018umap,mcqueen1967some,ward1963hierarchical}.
Representative samples are summarized into skill cards that describe
information structures, region patterns, density budgets, binding
patterns, and style constraints. At inference time, retrieved cards
guide IVD compilation, while evidence coverage and symbolic readability
take precedence over aesthetic preferences. Bank construction and
retrieval details are provided in the supplementary material.

\subsection{IVD as an Executable Contract}
\label{sec:ivd}

\paragraph{Typed element graph.}
IVD is represented as \(z=(\mathcal{E},\mathcal{U},\mathcal{R},G,\Omega)\), where \(\mathcal{E}\) is the evidence pool, \(\mathcal{U}\) defines canvas regions, \(\mathcal{R}\) specifies reading order, \(\Omega\) contains executable obligations, and \(G=(\mathcal{A},\mathcal{D})\) is a typed element dependency graph. Each element \(a_i\in\mathcal{A}\) follows the compact schema \(a_i=(m_i,p_i,r_i,b_i,c_i)\), where \(m_i\) stores its identifier, type, role, and priority; \(p_i\) contains an exact or paraphrasable payload together with its supporting evidence spans; \(r_i\) specifies its region and execution route; \(b_i\) records local binding relations; and \(c_i\subseteq\Omega\) indexes the obligations applied to the element. Typed edges in \(\mathcal{D}\) encode \texttt{contains}, \texttt{precedes}, \texttt{supports}, \texttt{binds-to}, and \texttt{depends-on} relations, allowing changes to an element to trigger the re-execution or re-checking of its affected dependents.

\paragraph{Executable obligations.}
Each obligation is represented as \(\omega_j=(S_j,\phi_j,\nu_j,\gamma_j)\), where \(S_j\) contains the scoped object--field identifiers to which the obligation applies, \(\phi_j\) is the required predicate, \(\nu_j\) is the checker, and \(\gamma_j\) denotes its severity. For example, a value element may require exact string equality, support from a cited evidence span, minimum contrast, and an arrow endpoint within a designated target region. Terms such as \texttt{exact}, \texttt{support}, and \texttt{no-overlap} therefore refer to checkable predicates with explicit scopes and repair targets, rather than descriptive tags.

\paragraph{Compilation and capability-constrained routing.}
The compiler decomposes the requested message into claims, values, dates, formulas, labels, citations, and visual intents. Exact strings are copied from evidence into protected payload fields, while paraphrasable claims retain their evidence spans. It then instantiates regions, reading order, element dependencies, local bindings, and executable obligations.

Each element is assigned to the smallest combination of layers capable of satisfying its obligations. Let \(\mathcal{L}=\{\mathrm{vis},\mathrm{sym},\mathrm{bind}\}\), let \(\operatorname{req}(a_i)\) denote the capabilities required by element \(a_i\), and define \(\operatorname{cap}(L)=\bigcup_{\ell\in L}\operatorname{cap}(\ell)\). The execution route is selected by
\begin{equation}
\begin{aligned}
r_i^\star
&=\arg\min_{\varnothing\neq L\subseteq\mathcal{L}}
\bigl(|L|,\operatorname{cost}(L)\bigr)\\
&\mathrm{s.t.}\quad
\operatorname{req}(a_i)\subseteq\operatorname{cap}(L),
\end{aligned}
\label{eq:routing}
\end{equation}

where the objective is minimized lexicographically, first by the number of layers and then by execution cost. Since \(\mathcal{L}\) contains only three layers, the implementation enumerates its seven non-empty subsets. Exact strings, numerical values, formulas, and vector geometry require symbolic execution; arrows, callouts, and object--label relations require binding execution; and open-ended appearance requires visual synthesis. Multiple layers are assigned only when no single layer covers all obligations. The capability matrix and execution-cost order are provided in the supplementary material.

\subsection{Trace-Preserving Layered Execution}
\label{sec:execution}

Each execution layer returns both a rendered artifact and its trace, while composition preserves their element-level or region-level associations:
\begin{equation}
\begin{aligned}
(I_\ell,T_\ell) &= R_\ell(z_\ell),\\
(y,T) &= \operatorname{Compose}\!\left(\{(I_\ell,T_\ell)\}_{\ell\in\mathcal{L}}\right),
\end{aligned}
\label{eq:execution}
\end{equation}
where \(z_\ell\) contains the IVD elements routed to layer \(\ell\), and \(T\) is the merged execution trace. Symbolic and binding elements retain exact rendering-tree handles, including their payloads, bounding boxes, z-order, and anchor geometry. Raster elements retain region-level traces from their assigned generation regions. When a binding or checker requires an object-level target, an independent grounding parser associates the corresponding IVD element with a box or mask and returns a confidence score; low-confidence associations are marked \textsc{Unknown}. Traces are therefore exact for editable vector objects and region- or grounding-based for raster content.

The \emph{visual layer} synthesizes backgrounds, visual metaphors, textures, and non-critical icons while preserving IVD-reserved symbolic regions. The \emph{symbolic layer} renders exact text, values, formulas, charts, legends, and source notes as editable SVG objects. The \emph{binding layer} renders arrows, leader lines, callouts, local labels, and legend--object mappings over resolved anchors. Its geometry is deterministic once the source and target anchors are established, while anchor resolution for raster objects may remain model-assisted. Further details on trace construction and anchor resolution are provided in the supplementary material.

\subsection{Verification and Scoped Repair}
\label{sec:repair}

\paragraph{Three-valued verification.}
For each obligation \(\omega_j\), the corresponding checker returns \(\nu_j(y,T,\mathcal{E})=(s_j,\kappa_j,w_j)\), where \(s_j\in\{\mathrm{PASS},\mathrm{FAIL},\mathrm{UNKNOWN}\}\) is its status, \(\kappa_j\) is the confidence, and \(w_j\) is a diagnostic witness. Rendering-tree checkers evaluate exact strings, values, formulas, clipping, overlap, contrast, z-order, and endpoint geometry. Evidence checkers compare claims with their cited spans, while visual-semantic checkers inspect only the object or crop identified by \(T\). Model-assisted checkers use fixed prompts and deterministic decoding, with pass and fail thresholds calibrated on a held-out human-labeled set; scores between the thresholds yield \textsc{Unknown}.

\(\mathcal{C}(y,z)=\{(\operatorname{id}(\omega_j),s_j,\kappa_j,w_j,\operatorname{ver}(\nu_j))\}_{\omega_j\in\Omega}\) is the resulting certificate. It records the status of each obligation together with the checker version and its witness. Witnesses may include cited evidence spans, expected and rendered strings, conflicting boxes, or unresolved target anchors. Obligations returning \textsc{Fail}, together with critical obligations returning \textsc{Unknown}, are emitted as element-indexed violations for subsequent scoped repair.

\paragraph{Transactional repair.}
A violation \(q=(\omega,s,\kappa,w)\) identifies a field and patch class. The dependency graph induces an impact closure \(H(q)\) containing elements that may need to be changed or re-checked. Candidate patches are drawn from a typed operator library, including evidence-backed payload replacement, symbolic re-rendering, constrained layout reflow, anchor reassignment, masked visual editing, and regional regeneration. The repair scope expands from field to element, region, and, when necessary, global recompilation.

\(\operatorname{Scope}(H)=\bigcup_{a_i\in H}\operatorname{scope}(a_i)\) collects the object--field identifiers affected by a repair. The protected set is \(P(H)=\{\omega:s(\omega)=\mathrm{PASS},\,S_\omega\cap\operatorname{Scope}(H)=\varnothing\}\), containing previously passed obligations outside this scope.

Let \(F^{\mathrm{crit}}\) and \(U^{\mathrm{crit}}\) denote the numbers of critical failures and critical unknowns, and let \(F\) and \(U\) denote their totals over all obligations. Candidate certificates are ranked by
\begin{equation}
Q^\star(\mathcal{C})=
\bigl(
F^{\mathrm{crit}}+U^{\mathrm{crit}},
F+U,
F^{\mathrm{crit}},
F
\bigr).
\label{eq:quality}
\end{equation}
A candidate is admissible only if the targeted critical obligation becomes \textsc{Pass}, or the total number of critical non-pass obligations strictly decreases. No other critical obligation may change from \textsc{Pass} to a non-pass state, and every obligation in \(P(H)\) must remain passed. Among admissible candidates, InfoAgent minimizes \(Q^\star\), followed by the number of changed elements, edited canvas area, and tool cost. If no candidate satisfies these conditions, the current candidate is rolled back and the repair scope is expanded.

This acceptance rule provides checker-relative preservation for previously passed obligations outside the affected dependency scope. It does not exclude changes within that scope or errors not captured by the declared checkers. Repair stops when all critical obligations pass, no admissible patch is found, or the maximum of \(K\) rounds is reached; unresolved obligations remain explicit in the returned certificate. The patch library, dependency propagation rules, and full repair algorithm are provided in the supplementary material.
\section{Experiments}
\label{sec:experiments}

\subsection{Experimental Setup}
\label{sec:exp_setup}

We evaluate atomic and whole-infographic reliability, executable IVD beyond prompt serialization, the effectiveness and locality of scoped repair, and the trade-offs among reliability, visual quality, output coverage, and cost.

\paragraph{Benchmarks.}
\textbf{IGenBench}~\cite{tang2026igenbench} contains 600 prompts, 30 infographic types, and 5,259 atomic questions from ten reliability categories. We use the benchmark under its official one-output protocol; the generation pipeline receives only the prompt and never accesses benchmark questions, category labels, evaluator prompts, or outputs. \textbf{InfoGraphicBench-Evidence} contains 260 requests across six knowledge domains, split into 30 development, 30 calibration, and 200 topic-disjoint test instances. Each instance provides a request and evidence bundle, with held-out annotations for factual and symbolic requirements and text--visual bindings. Test instances contain 8.4 requirements, 4.7 required text units, 2.3 numerical or source-note units, and 2.6 bindings on average. Construction, deduplication, and annotation details appear in the supplementary material.

\paragraph{Baselines.}
\textbf{Direct T2I} uses Qwen-Image~\cite{wu2025qwen} to render the original request without external evidence, planning, verification, or repair. \textbf{RAG Prompt}, used only on InfoGraphicBench-Evidence, serializes the shared evidence into a generation prompt; \textbf{Same-IVD Prompt} instead serializes the automatically compiled IVD before verification or repair and delegates complete rendering to the same image generator. \textbf{Gen-Searcher}~\cite{feng2026gen} and \textbf{GenClaw}~\cite{ye2026genclaw} are rerun under our controlled protocol while retaining their native search/planning and code-driven execution strategies, respectively. \textbf{LLM-to-SVG/HTML} renders the complete infographic as executable vector or web code using the same language-model backbone. \textbf{InfoAgent w/o repair} is an internal \(K=0\) variant that retains IVD compilation and layered execution but applies no verifier-triggered patch. Benchmark-reported IGenBench rows are shown only as external reference points; adaptation and execution details are provided in the supplementary material.

\begin{table}[t]
\centering
\caption{Reliability on the complete IGenBench benchmark.}
\resizebox{0.7\linewidth}{!}{
\begin{tabular}{@{}lccc@{}}
\toprule
Method & Q-ACC $\uparrow$ & I-ACC $\uparrow$ & Out $\uparrow$ \\
\midrule
\multicolumn{4}{@{}l}{\emph{Benchmark-reported references~\cite{tang2026igenbench}}} \\
NanoBanana-Pro~\cite{nanobanana-pro}
& 90.0 & 49.0 & -- \\
Seedream-4.5~\cite{seedream4.5}
& 61.0 & 6.0 & -- \\
GPT-Image-1.5~\cite{chatgpt-images}
& 55.0 & 12.0 & -- \\
\midrule
\multicolumn{4}{@{}l}{\emph{Controlled baselines}} \\
Direct T2I~\cite{wu2025qwen}
& 43.0 & 2.0 & 100.0 \\
Same-IVD Prompt
& 50.0 & 5.0 & 100.0 \\
Gen-Searcher~\cite{feng2026gen}
& 76.0 & 20.0 & 100.0 \\
GenClaw~\cite{ye2026genclaw}
& 79.0 & 27.0 & 98.2 \\
LLM-to-SVG/HTML
& 82.0 & 32.0 & 96.3 \\
\midrule
InfoAgent w/o repair
& 88.0 & 45.0 & 100.0 \\
\textbf{InfoAgent}
& \textbf{93.0} & \textbf{59.0} & 100.0 \\
\bottomrule
\end{tabular}
}
\label{tab:igenbench_main}
\end{table}

\paragraph{Metrics.}
On IGenBench, \textbf{Q-ACC} averages atomic-question accuracy, while \textbf{I-ACC} requires every question for an infographic to be correct. On InfoGraphicBench-Evidence, \textbf{Out} measures valid-output coverage, \textbf{ReqCov} averages requirement satisfaction, and \textbf{Full} requires the external checklist to pass without additional critical errors. \textbf{NonSup.} is the percentage of factual claims unsupported or contradicted by evidence; \textbf{ReqText-F1} matches required and recognized strings character-wise; \textbf{Bind-Q} rates local text--visual grounding on a 1--5 scale; and \textbf{Q-Align}~\cite{wu2023q} measures perceptual quality. \cref{tab:mechanism_ablation} retains the latter three metrics and adds mechanism-sensitive endpoints: \textbf{CritPass} requires all critical factual, symbolic, and binding obligations to pass, \textbf{Fact} counts omitted required facts as failures, and \textbf{Layout-Q} rates region organization, reading flow, hierarchy, overlap, and crowding. Formal definitions are provided in the supplementary material.

\paragraph{Protocol and implementation.}
All evidence-conditioned methods receive the same request and evidence bundle, while Direct T2I receives only the request. Where applicable, methods use the same Qwen-Image renderer~\cite{wu2025qwen}, canvas resolution, and visual-reference budget; native generation and code-rendering pipelines retain their original renderers. Our primary configuration uses Gemini 3.1 Pro~\cite{team2026gemini} for evidence organization, design-prior selection, IVD compilation, and patch proposal. Claude Opus 4.6~\cite{claude-opus-4.6} is evaluated as an alternative proprietary planner backbone in the supplementary material. Rendering-tree checks are deterministic; model-assisted online verification uses a separate scoped verifier; and the offline evaluator receives final outputs and held-out annotations but no IVD fields, execution traces, repair logs, or checker witnesses.

All evaluation annotations remain hidden from generation and repair. Main results use one output per request and at most \(K=3\) repair rounds. Checker and repair analysis uses 600 stratified obligation decisions from 120 fully audited infographics, while certificate calibration uses all 200 test outputs. We report paired-bootstrap 95\% confidence intervals from 10,000 resamples and Holm-adjusted \(p\)-values for pre-specified primary tests. Invalid outputs receive zero for ReqCov, Full, ReqText-F1, and end-to-end Bind-Q; NonSup. and Q-Align use valid outputs, with common-valid Q-Align, model snapshots, prompts, hardware, robustness, retrieval, and cost analyses reported in the supplementary material.

\begin{table*}[t]
\centering
\caption{Controlled results on 200 InfoGraphicBench-Evidence test requests. Evidence-conditioned methods use the same evidence bundle; Direct T2I is evidence-free.}
\resizebox{0.916\linewidth}{!}{
\begin{tabular}{@{}lrrrrrrr@{}}
\toprule
Method & Out $\uparrow$ & ReqCov $\uparrow$ & Full $\uparrow$ & NonSup. (\%) $\downarrow$ & ReqText-F1 $\uparrow$ & Bind-Q $\uparrow$ & Q-Align $\uparrow$ \\
\midrule
Direct T2I~\cite{wu2025qwen}
& 100.0 & 76.74 & 17.5 & 1.63 & 82.7 & 4.53 & 4.508 \\
RAG Prompt
& 100.0 & 79.40 & 19.5 & 1.18 & 83.3 & 4.56 & 4.515 \\
Same-IVD Prompt
& 100.0 & 81.00 & 21.5 & 1.02 & 86.4 & 4.58 & 4.522 \\
Gen-Searcher~\cite{feng2026gen}
& 100.0 & 82.10 & 22.5 & 0.87 & 85.4 & 4.61 & 4.556 \\
GenClaw~\cite{ye2026genclaw}
& 98.0 & 81.50 & 25.0 & 0.76 & 92.7 & 4.60 & 4.048 \\
LLM-to-SVG/HTML
& 96.5 & 79.61 & 23.0 & 0.80 & 94.7 & 4.58 & 3.920 \\
\midrule
InfoAgent w/o repair & 100.0 & 81.92 & 23.5 & 0.89 & 92.4 & 4.47 & 4.502 \\
\textbf{InfoAgent}
& 100.0 & \textbf{83.54} & \textbf{28.5} & \textbf{0.57} & \textbf{96.5} & \textbf{4.64} & \textbf{4.621} \\
\bottomrule
\end{tabular}
}
\label{tab:evidence_main}
\end{table*}
\begin{table*}[t]
\centering
\caption{Ablations on InfoGraphicBench-Evidence. ReqText-F1, Bind-Q, and Q-Align are shared with~\cref{tab:evidence_main}; CritPass, Fact, and Layout-Q isolate critical reliability, required-fact preservation, and layout quality.}
\resizebox{0.916\linewidth}{!}{
\begin{tabular}{@{}lrrrrrr@{}}
\toprule
Variant & CritPass $\uparrow$ & Fact $\uparrow$ & Layout-Q $\uparrow$ & ReqText-F1 $\uparrow$ & Bind-Q $\uparrow$ & Q-Align $\uparrow$ \\
\midrule
\textbf{Full InfoAgent} & \textbf{61.5} & \textbf{91.6} & \textbf{4.45} & \textbf{96.5} & \textbf{4.64} & \textbf{4.621} \\

InfoAgent w/o repair & 46.0 & 88.0 & 4.30 & 92.4 & 4.47 & 4.502 \\

\midrule

Flat plan & 42.5 & 86.8 & 4.13 & 88.1 & 4.24 & 4.389 \\

Symbolic content to raster & 38.5 & 90.1 & 4.43 & 85.2 & 4.55 & 4.529 \\

w/o evidence binding & 49.0 & 84.2 & 4.44 & 92.9 & 4.58 & 4.515 \\

w/o binding route & 45.0 & 89.9 & 4.42 & 93.0 & 4.19 & 4.511 \\

w/o design-prior retrieval & 54.0 & 89.8 & 4.18 & 93.1 & 4.53 & 4.407 \\

Whole-image feedback & 48.5 & 88.7 & 4.29 & 91.4 & 4.41 & 4.487 \\

w/o execution trace & 44.0 & 88.9 & 4.25 & 89.8 & 4.27 & 4.476 \\
\bottomrule
\end{tabular}

}
\label{tab:mechanism_ablation}
\end{table*}
\begin{table}[t]
\centering
\scriptsize
\caption{Repair effectiveness and locality on 120 audited repair-eligible outputs.}
\resizebox{0.8\linewidth}{!}{
\begin{tabular}{@{}lrrrrr@{}}
\toprule
Strategy & \shortstack{Fix@\\Det} & \shortstack{Repair\\Rec.} & \shortstack{New\\Img.} & \shortstack{Collat.\\Reg.} & \shortstack{Area\\Edit} \\
\midrule
\textbf{Localized repair} & \textbf{85.9} & \textbf{77.8} & \textbf{3.3} & \textbf{0.8} & 12.4 \\
w/o dependency closure & 79.1 & 71.7 & 7.5 & 4.9 & \textbf{7.8} \\
Global regeneration & 74.8 & 67.8 & 10.0 & 6.8 & 67.3 \\
\bottomrule
\end{tabular}
}
\vspace{-4mm}
\label{tab:repair_locality}
\end{table}

\subsection{Reliability on IGenBench}
\label{sec:igenbench_results}

\cref{tab:igenbench_main} separates benchmark-reported references, controlled baselines, and InfoAgent variants. InfoAgent achieves 93.0 Q-ACC and 59.0 I-ACC, exceeding the strongest benchmark-reported reference by 10.0 I-ACC points. Among controlled runs, search- and code-oriented baselines improve over Direct T2I, while Same-IVD Prompt remains well below InfoAgent w/o repair, showing that prompt serialization does not reproduce executable layered rendering. Scoped repair further raises I-ACC from 45.0 to 59.0, a paired gain of 14.0 points (95\% CI [10.4, 17.6], Holm-adjusted \(p<0.001\)).

\begin{figure*}[t]
\centering
\includegraphics[width=\linewidth]{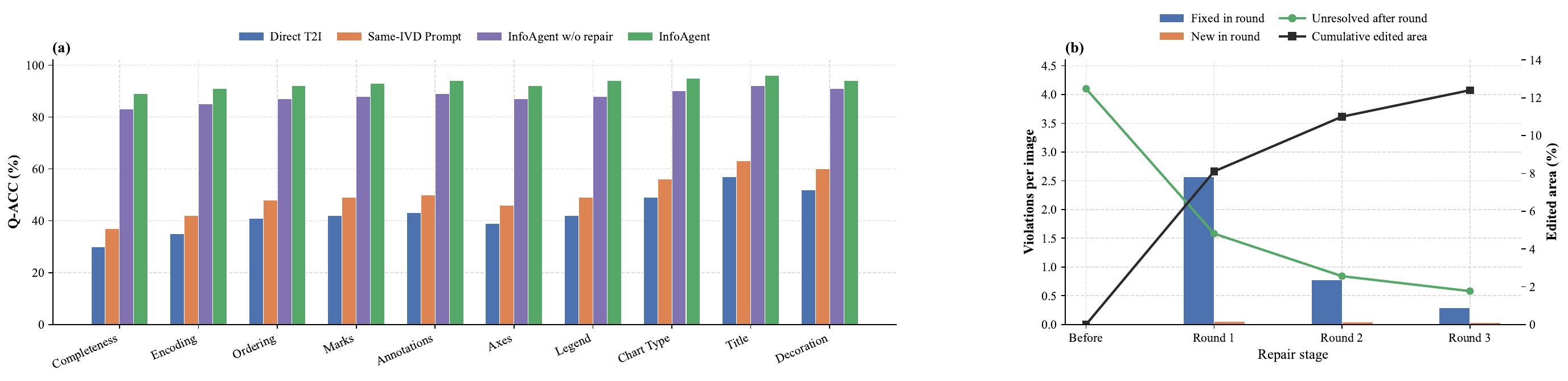}
\caption{Reliability breakdown and scoped-repair dynamics. (a) Category-wise Q-ACC on IGenBench. (b) Mean detected critical violations and cumulative edited area over three repair rounds on 120 audited repair-eligible outputs.}
\label{fig:igenbench_repair}
\end{figure*}

\cref{fig:igenbench_repair}(a) shows gains across the ten IGenBench categories, with the largest improvements in completeness, encoding, marks, axes, annotations, and legend mapping. \cref{fig:igenbench_repair}(b) shows that most detected critical violations are resolved within the first two rounds, while the third round yields smaller gains with limited additional edited area.

\subsection{Evidence-Controlled Evaluation}
\label{sec:evidence_results}

\cref{tab:evidence_main} shows a clear progression from evidence conditioning to structured planning, executable rendering, and repair. RAG Prompt improves over evidence-free Direct T2I, while Same-IVD Prompt provides further gains. InfoAgent w/o repair raises Full from 21.5\% to 23.5\% and ReqText-F1 from 86.4 to 92.4 over Same-IVD Prompt, showing that layered execution contributes beyond prompt serialization; scoped repair further increases them to 28.5\% and 96.5. Code-oriented baselines achieve strong text fidelity but lower perceptual quality. Full InfoAgent retains 100\% output coverage and obtains the best point estimates on all reported endpoints. Its Full gain over GenClaw is 3.5 points (95\% CI [0.6, 6.4], Holm-adjusted \(p=0.041\)), while its ReqText-F1 gain over LLM-to-SVG/HTML is 1.8 points (95\% CI [0.8, 2.8], adjusted \(p=0.006\)).

\subsection{Mechanism and Repair Analysis}
\label{sec:mechanism}

We isolate addressability through targeted interventions. \textbf{Flat plan} retains compiled payloads and coarse layout but removes typed identifiers, dependency edges, executable obligations, and traces. \textbf{Whole-image feedback} replaces element-indexed violations with holistic critique and global refinement, while \textbf{w/o execution trace} requires post-hoc relocation of rendered elements. Full definitions are provided in the supplementary material.

\cref{tab:mechanism_ablation} shows that structured planning is insufficient. Flat plan reduces CritPass from 61.5\% to 42.5\%. Routing symbolic content through raster generation causes the largest ReqText-F1 drop, from 96.5 to 85.2; removing the binding route reduces Bind-Q, from 4.64 to 4.19; and removing design-prior retrieval most strongly affects Layout-Q and Q-Align. Without repair, CritPass is 46.0\%. Dependency-aware repair raises it to 61.5\%, compared with 54.5\% without dependency closure and 50.5\% under global regeneration. As shown in \cref{tab:repair_locality}, the verifier detects 90.6\% of confirmed critical violations, and localized repair fixes 85.9\% of detected violations, yielding 77.8\% end-to-end recall. It edits 12.4\% of the canvas on average, versus 67.3\% for global regeneration, while producing fewer new and collateral errors.

\paragraph{Verifier and certificate audit.}
Checker performance is evaluated on 600 obligation decisions from 120 audited infographics, with thresholds fixed on the separate calibration split. Rendering-tree, evidence-support, binding, and visual-semantic checks obtain 96.3, 88.1, 84.9, and 82.7 F1, respectively; human agreement reaches Fleiss' \(\kappa=0.82\). On all 200 test outputs, internal certification achieves 92.2\% precision and 86.2\% recall against external CritPass labels. Detailed counts, \textsc{Unknown} rates, and confusion matrices are provided in the supplementary material.

\subsection{Qualitative Comparison}
\label{sec:qualitative}

\begin{figure*}[t]
\centering
\includegraphics[width=\linewidth]{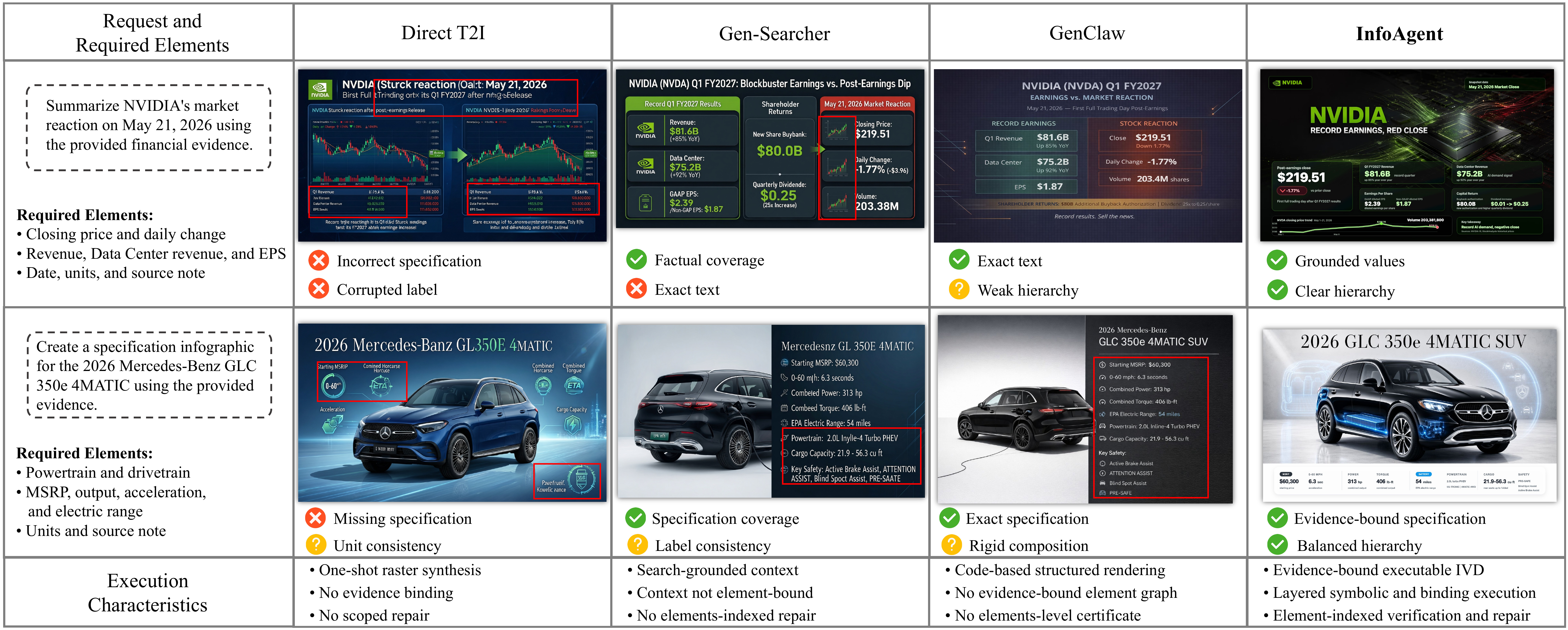}
\caption{Qualitative comparison on financial-event and product-specification requests. Boxes mark factual, symbolic, binding, and hierarchy differences; \(\checkmark\), \(\times\), and \(?\) denote satisfied, violated, and partially satisfied requirements. All outputs are shown without manual correction.}
\label{fig:evidence_qualitative}
\vspace{-3mm}
\end{figure*}

\cref{fig:evidence_qualitative} compares a time-sensitive financial request with a specification-intensive product request. Direct T2I produces visually plausible compositions, but may omit required specifications, corrupt labels, or render unsupported values. Gen-Searcher improves factual and specification coverage through external context, while exact text and label consistency remain less stable because the retrieved evidence is used primarily at the prompt level. GenClaw provides more reliable symbolic content through code-based structured rendering, but its outputs exhibit weaker visual hierarchy and more rigid compositions. InfoAgent instead compiles evidence into addressable IVD elements and executes factual values, specifications, and local relations through coordinated visual, symbolic, and binding layers. The annotated boxes and status markers highlight differences in factual accuracy, symbolic fidelity, label consistency, and visual hierarchy. Additional randomly sampled outputs, failure cases, and repair trajectories are provided in the supplementary material.

\paragraph{Limitations.}
InfoAgent improves strict reliability but does not make every output correct. On IGenBench, 41.0\% of outputs still fail at least one atomic question. On InfoGraphicBench-Evidence, 28.5\% pass the complete external checklist, while 61.5\% pass all critical factual, symbolic, and binding requirements. Common remaining failures include incomplete or conflicting evidence, uncertain grounding of raster objects, and global layout errors that often require broader replanning. Model-assisted evidence and binding checks also remain less reliable than rendering-tree checks; uncertain obligations are therefore retained as \textsc{Unknown} rather than treated as certified.
\section{Conclusion}

We presented \textbf{InfoAgent}, a training-free framework that links evidence, rendered elements, and verification obligations through IVD. Layered execution retains traces for locating errors, while dependency-aware repair localizes corrections and rechecks affected dependencies and protected obligations. With evidence and the initial IVD held fixed, the complete-checklist pass rate on InfoGraphicBench-Evidence increases from 21.5\% for Same-IVD Prompt to 28.5\% for InfoAgent. Together with the IGenBench results and repair-locality analysis, these findings support preserving evidence links and execution dependencies throughout generation and revision.

Incomplete or conflicting evidence, uncertain raster-object grounding, and imperfect semantic checks remain sources of failure. The certificate records outcomes under the declared obligations and checkers, retaining unresolved requirements rather than treating them as satisfied. Improving evidence reconciliation and visual grounding, and extending IVD to interactive and multi-page artifacts, are directions for future work.
\section{AI Use Statement}

We used Gemini (Google) only for grammar correction and language polishing during the writing of this paper. No AI tools were involved in the research ideation, methodology, experiments, or analysis. The authors take full responsibility for all content in this manuscript.

\bibliography{preprint}
\bibliographystyle{preprint}

\renewcommand{\thesection}{\Alph{section}}
\renewcommand{\thesubsection}{\thesection.\arabic{subsection}}
\renewcommand{\thetable}{S\arabic{table}}
\renewcommand{\thefigure}{S\arabic{figure}}
\renewcommand{\theequation}{S\arabic{equation}}
\setcounter{table}{0}
\setcounter{figure}{0}
\setcounter{equation}{0}

\providecommand{\mcpdeqref}[1]{(\ref{#1})}
\clearpage
\appendix
\ifdefined\startcontents\startcontents[supp]\fi
\begin{center}
{\LARGE\bfseries Supplementary Material}
\end{center}
\ifdefined\printcontents
\medskip
{\large\textsc{Contents}}
\printcontents[supp]{}{1}{}
\clearpage
\fi

\section{Benchmark and Evaluation Details}
\label{sec:supp_benchmark}

\subsection{InfoGraphicBench-Evidence}
\label{sec:supp_benchmark_construction}

InfoGraphicBench-Evidence contains 260 knowledge-intensive infographic requests spanning public health, education, environment, economy, social trends, and science communication. We use 30 requests for development, 30 for checker and evaluator calibration, and 200 topic-disjoint requests for testing. Each instance contains a user request and a frozen evidence bundle; atomic requirements, required strings, factual items, numerical or source-note units, text--visual bindings, and evaluator questions are held out for evaluation. Test requests contain 8.4 atomic requirements, 4.7 required text units, 2.3 numerical or source-note units, and 2.6 bindings on average.

\begin{table*}[t]
\centering
\caption{InfoGraphicBench-Evidence domain distribution.}
\resizebox{0.8\linewidth}{!}{
\begin{tabular}{@{}lrrrrrrr@{}}
\toprule
Split & Public health & Education & Environment & Economy & Social trends & Science communication & Total \\
\midrule
Development & 5 & 5 & 5 & 5 & 5 & 5 & 30 \\
Calibration & 5 & 5 & 5 & 5 & 5 & 5 & 30 \\
Test & 34 & 33 & 34 & 33 & 33 & 33 & 200 \\
\midrule
Total & 44 & 43 & 44 & 43 & 43 & 43 & 260 \\
\bottomrule
\end{tabular}
}
\label{tab:supp_domain_distribution}
\end{table*}

\paragraph{Sources and topic separation.}
Evidence bundles were assembled from authoritative public sources and frozen before generation. Each factual span retains a stable evidence identifier, publisher, source title, URL, publication and access dates, document hash, and local span offsets. Requests were grouped by normalized entities and event templates and then filtered using BGE-M3 semantic embeddings with cosine similarity above 0.82; candidate overlaps were manually reviewed, and related topics were assigned to the same split.

\paragraph{Annotation and release.}
Two annotators independently construct the factual items, required strings, bindings, and atomic checklist from the frozen evidence. A third senior annotator adjudicates disagreements. An item is critical when its omission or corruption can invalidate the central claim, an exact value or source attribution, or an essential text--visual relation; decorative details and secondary layout preferences are non-critical. The release contains requests, source metadata, evidence identifiers and offsets, checklists, bindings, split manifests, and evaluation scripts. Source text is redistributed only when its license permits; otherwise, the release provides URLs, hashes, and offsets rather than copyrighted documents.

\paragraph{Leakage control.}
Evidence-conditioned methods receive only the request and shared evidence bundle. Checklists, required-text lists, binding annotations, evaluator questions, and external critical-error labels are never exposed to baselines, the IVD compiler, renderers, online verifier, or repair procedure. On IGenBench, generation receives only the benchmark prompt and never accesses atomic questions, category labels, evaluator prompts, or evaluator outputs.

\subsection{Metric Definitions}
\label{sec:supp_metrics}

Let \(N\) be the number of requests and \(V_n\) indicate a valid output. \textbf{Out} is \(100N^{-1}\sum_n V_n\). For request \(n\), let \(\mathcal{Q}_n\) be its external checklist and \(q_{nj}\in\{0,1\}\) the evaluated result. \textbf{ReqCov} averages \( |\mathcal{Q}_n|^{-1}\sum_j q_{nj}\) over all requests, assigning zero to invalid outputs. \textbf{Full} is the percentage of requests for which every checklist item passes and no additional critical factual, symbolic, binding, or layout error is present. \textbf{CritPass} instead requires all critical factual, symbolic, and binding items to pass and excludes non-critical decorative and secondary layout requirements.

\textbf{NonSup.} is the percentage of factual claims visibly present in valid outputs that are unsupported or contradicted by the shared evidence. It is not the complement of \textbf{Fact}: Fact evaluates benchmark-required factual items and counts omissions as failures. \textbf{Bind-Q} is a 1--5 external-evaluator rating of label, arrow, callout, legend, and annotation grounding. \textbf{Layout-Q} rates region organization, reading flow, hierarchy, alignment, overlap, boundary compliance, and crowding on the same scale. \textbf{Q-Align} measures perceptual quality.

\paragraph{Required-text matching.}
Recognized text regions are evaluated outside the generation and repair pipeline. We apply Unicode NFKC normalization, case folding where appropriate, whitespace and punctuation normalization, and canonical formatting for numbers, percentages, dates, and units. Required units and recognized regions are aligned one-to-one by maximum-weight bipartite matching. Exact values, formulas, source identifiers, and dates use strict or near-strict matching; ordinary phrases allow a bounded character-edit tolerance. We report character-weighted precision, recall, and their harmonic mean, \textbf{ReqText-F1}. Invalid outputs receive zero.

\paragraph{Repair metrics.}
\textbf{DetRec} is the fraction of externally confirmed critical violations detected by the online verifier. \textbf{Fix@Det} is the fraction of correctly detected true violations fixed by repair, and \textbf{RepairRec} is the fraction of all externally confirmed violations both detected and fixed. \textbf{NewImg} is the percentage of repaired outputs containing at least one newly introduced externally confirmed violation. \textbf{CollatReg} measures regression among previously passed obligations outside the target violation's original dependency closure. \textbf{AreaEdit} is the union area of committed canvas patches divided by canvas area; overlapping edits are counted once.

\paragraph{Statistics.}
We use request-level paired bootstrap 95\% confidence intervals from 10,000 resamples. Binary image-level comparisons use exact McNemar tests, continuous endpoints use paired randomization tests, and pre-specified primary \(p\)-values are adjusted by Holm's procedure.

\section{Implementation, Baselines, and Compute}
\label{sec:supp_implementation}

\subsection{System and Comparison Protocol}
\label{sec:supp_protocol}

The primary configuration uses Gemini 3.1 Pro for evidence organization, design-prior selection, IVD compilation, and patch proposal. Qwen-Image is used as the shared raster renderer when a baseline permits it. Rendering-tree checks are deterministic. The scoped semantic verifier and primary offline evaluator use separate Gemini invocations, prompts, and visible inputs. This provides procedural isolation but not model-family independence. The offline evaluator receives final outputs and held-out annotations but no IVD fields, execution traces, repair logs, checker witnesses, or generator identities.

\paragraph{Baseline adapters.}
Direct T2I receives only the request. RAG Prompt serializes the shared evidence, while Same-IVD Prompt serializes the automatically compiled pre-repair IVD and delegates complete rendering to Qwen-Image. Gen-Searcher and GenClaw retain their native planning and rendering strategies under the shared evidence protocol. LLM-to-SVG/HTML uses the same planner backbone to synthesize executable vector or web code. InfoAgent without repair retains the complete IVD and layered executor but sets \(K=0\). Invalid generations, unparsable code, and rendering failures count as invalid outputs.

\paragraph{Contemporary agentic baselines.}
The budget-matched study additionally includes M3, the closest plan--check--edit baseline, and Qwen-Image-Agent, a context-centric search and feedback agent. We do not report empirical results for GenEvolve or Unify-Agent, because our evaluation did not obtain a controlled adapter that preserved their defining jointly trained components while fixing the evidence bundle and renderer. Accordingly, empirical superiority claims are restricted to the methods actually executed.

\subsection{Inference Cost and Cost-Capped Evaluation}
\label{sec:supp_compute}

We first report the native inference profile of each method and then evaluate reliability under several online-cost caps. The two analyses serve different purposes: the native profile describes the resources used by each original pipeline, whereas the cost-capped study examines whether the reliability advantage persists when long trajectories are truncated.

\paragraph{Accounting protocol.}
Call categories are mutually exclusive. \textbf{Plan} counts all non-verifier language-model calls, including initial planning, IVD or code compilation, and repair-patch proposal. \textbf{Search} counts online textual or visual search APIs. \textbf{Raster} counts image generation and image-editing calls. \textbf{Verify} counts model-assisted online verification and re-verification calls. Thus, a repair may contribute one Plan call, one Raster call, and several Verify calls, but it is not counted again as a separate event. Deterministic rendering-tree checks and local SVG execution do not consume model calls.

Tokens sum uncached text input and output tokens over planning, verification, and repair. Image inputs and outputs are billed using the corresponding provider image prices and are not converted into text-token equivalents. Search charges are included in Cost but not Tokens. Reported cost uses the fixed provider price sheet archived with the released experiment manifest; cached-token discounts are ignored. Latency is sequential wall-clock time from the first online call to the final artifact, including API communication, failed-call retries, raster rendering, SVG composition, and verification. We report mean and standard deviation over all 200 requests. Provider-side accelerator time and memory are unavailable and are not inferred. Local SVG rendering uses a 32-core CPU and averages 1.9\,s with 1.4\,GB peak RAM per InfoAgent output.

\begin{table*}[t]
\centering
\caption{Native online inference profile on InfoGraphicBench-Evidence. Calls are mutually exclusive averages per request.}
\begin{tabular}{@{}lrrrrrrr@{}}
\toprule
Method
& Plan
& Search
& Raster
& Verify
& Tokens (k)
& Latency (s)
& Cost (\$) \\
\midrule
Direct T2I
& 0.00 & 0.00 & 1.00 & 0.00
& 0.6 & 17.8\(\pm\)4.6 & 0.040 \\

Same-IVD Prompt
& 1.00 & 0.00 & 1.00 & 0.00
& 7.1 & 31.6\(\pm\)8.4 & 0.084 \\

Gen-Searcher
& 2.75 & 1.42 & 1.35 & 0.90
& 12.8 & 78.4\(\pm\)23.1 & 0.191 \\

GenClaw
& 3.43 & 0.00 & 0.82 & 1.55
& 15.1 & 91.3\(\pm\)28.7 & 0.176 \\

M3
& 4.40 & 0.00 & 1.75 & 2.40
& 16.2 & 96.8\(\pm\)31.5 & 0.213 \\

Qwen-Image-Agent
& 3.65 & 1.18 & 1.55 & 1.75
& 14.7 & 88.5\(\pm\)27.4 & 0.201 \\

LLM-to-SVG/HTML
& 3.05 & 0.00 & 0.00 & 1.30
& 18.6 & 44.2\(\pm\)15.9 & 0.133 \\

InfoAgent (\(K=0\))
& 1.00 & 0.00 & 1.00 & 0.00
& 9.4 & 38.1\(\pm\)9.6 & 0.102 \\

InfoAgent
& 2.62 & 0.00 & 1.18 & 3.06
& 14.8 & 71.9\(\pm\)24.8 & 0.168 \\
\bottomrule
\end{tabular}
\label{tab:supp_native_compute}
\end{table*}

InfoAgent incurs additional verification and repair cost relative to its \(K=0\) output, but remains less expensive on average than Gen-Searcher, GenClaw, M3, and Qwen-Image-Agent under their native pipelines. The comparison is descriptive and does not equate hidden provider compute.

\paragraph{Cost-capped evaluation.}
Rather than selecting a single threshold around InfoAgent's native cost, we use four caps fixed before analysis: \$0.10, \$0.15, \$0.20, and \$0.25 per request. The curve is reconstructed from the same logged trajectories used for the main evaluation. For each request, execution stops before the first call that would exceed the cap and returns the latest valid artifact; if no valid artifact exists, the request is counted as an invalid output. Methods are not rerun or re-optimized for individual caps. Consequently, once the complete trajectory fits within a cap, its result matches the corresponding main-table result exactly.

\begin{table}[t]
\centering
\caption{Full under increasing online-cost caps. Values are percentages over the same 200 requests.}
\resizebox{\linewidth}{!}{
\begin{tabular}{@{}lrrrrr@{}}
\toprule
Method
& Native cost
& \$0.10
& \$0.15
& \$0.20
& \$0.25 \\
\midrule
Direct T2I
& 0.040 & 17.5 & 17.5 & 17.5 & 17.5 \\

Same-IVD Prompt
& 0.084 & 21.5 & 21.5 & 21.5 & 21.5 \\

Gen-Searcher
& 0.191 & 19.5 & 21.5 & 22.5 & 22.5 \\

GenClaw
& 0.176 & 20.5 & 23.5 & 25.0 & 25.0 \\

M3
& 0.213 & 20.5 & 23.5 & 25.5 & 26.0 \\

Qwen-Image-Agent
& 0.201 & 19.5 & 22.0 & 24.0 & 24.5 \\

LLM-to-SVG/HTML
& 0.133 & 21.0 & 23.0 & 23.0 & 23.0 \\

\textbf{InfoAgent}
& 0.168
& \textbf{23.0}
& \textbf{26.5}
& \textbf{28.5}
& \textbf{28.5} \\
\bottomrule
\end{tabular}
}
\label{tab:supp_cost_curve}
\end{table}

The cost curve distinguishes early executable output from later verification and repair gains. At \$0.10, InfoAgent operates close to its \(K=0\) configuration; additional budget permits scoped checking and patching, raising Full to 28.5\%. The method remains strongest at every tested cap, while the saturated values reproduce the main results. This experiment controls observable online monetary expenditure, not unreported provider-side computation.

\section{Additional Experiments}
\label{sec:supp_additional_experiments}

\subsection{Statistical Comparisons}
\label{sec:supp_statistics}

InfoAgent improves I-ACC over its \(K=0\) variant by 14.0 points (95\% CI \([10.4,17.6]\), Holm-adjusted \(p<0.001\)). On InfoGraphicBench-Evidence, the Full gain over GenClaw is 3.5 points (95\% CI \([0.6,6.4]\), adjusted \(p=0.041\)), and the ReqText-F1 gain over LLM-to-SVG/HTML is 1.8 points (95\% CI \([0.8,2.8]\), adjusted \(p=0.006\)). The Q-Align difference over Gen-Searcher is 0.065 (95\% CI \([0.018,0.112]\)); the Bind-Q difference is 0.03 (95\% CI \([-0.01,0.07]\)) and is treated as comparable.

\subsection{Robustness Diagnostics}
\label{sec:supp_robustness}

The following studies use 80 requests fixed before generation and stratified by domain, infographic type, requirement-count bin, and binding-count bin with seed 2027. They diagnose acquisition and sampling sensitivity rather than replace the complete 200-request evaluation.

\paragraph{Evidence acquisition and conflict.}
We compare the frozen evidence bundle with end-to-end retrieval and a conflicting candidate pool containing temporally inconsistent or definition-mismatched sources. The latter requires the planner to select, reconcile, or qualify evidence before IVD compilation.

\begin{table}[t]
\centering
\caption{Evidence robustness on 80 fixed requests. Brackets report request-level bootstrap 95\% confidence intervals.}
\resizebox{\linewidth}{!}{
\begin{tabular}{@{}lccccc@{}}
\toprule
Evidence setting
& ReqCov
& Full
& NonSup.
& ReqText-F1
& Bind-Q \\
\midrule
Frozen bundle
& \shortstack{84.1\\{\scriptsize [82.3,85.8]}}
& \shortstack{30.0\\{\scriptsize [21.1,40.8]}}
& \shortstack{0.55\\{\scriptsize [0.38,0.77]}}
& \shortstack{96.7\\{\scriptsize [96.0,97.3]}}
& \shortstack{4.65\\{\scriptsize [4.60,4.70]}} \\

Retrieved evidence
& \shortstack{81.2\\{\scriptsize [79.1,83.2]}}
& \shortstack{25.0\\{\scriptsize [16.8,35.5]}}
& \shortstack{1.08\\{\scriptsize [0.78,1.42]}}
& \shortstack{95.1\\{\scriptsize [94.2,95.9]}}
& \shortstack{4.60\\{\scriptsize [4.54,4.66]}} \\

Conflicting pool
& \shortstack{77.8\\{\scriptsize [75.4,80.1]}}
& \shortstack{20.0\\{\scriptsize [12.7,30.0]}}
& \shortstack{1.76\\{\scriptsize [1.32,2.25]}}
& \shortstack{93.8\\{\scriptsize [92.7,94.7]}}
& \shortstack{4.51\\{\scriptsize [4.44,4.58]}} \\
\bottomrule
\end{tabular}
}
\label{tab:supp_evidence_robustness}
\end{table}

The frozen-bundle setting provides an execution-focused reference. End-to-end retrieval mainly reduces factual coverage and strict pass rate, while conflicting evidence further increases unsupported claims. Required-text fidelity degrades more moderately because selected factual payloads are rendered through the same symbolic executor.

\paragraph{Repeated generation.}
We independently repeat the complete generation process three times on the same 80 requests while keeping prompts, evidence bundles, annotations, and method budgets fixed.

\begin{table}[t]
\centering
\caption{Three-generation robustness. Values are mean \(\pm\) standard deviation.}
\resizebox{\linewidth}{!}{
\begin{tabular}{@{}lrrrr@{}}
\toprule
Method & Full & ReqText-F1 & Bind-Q & Q-Align \\
\midrule
Same-IVD
& 21.3\(\pm\)1.3
& 86.4\(\pm\)0.6
& 4.57\(\pm\)0.03
& 4.52\(\pm\)0.02 \\

GenClaw
& 24.6\(\pm\)0.7
& 92.6\(\pm\)0.5
& 4.59\(\pm\)0.02
& 4.05\(\pm\)0.04 \\

InfoAgent (\(K=0\))
& 23.3\(\pm\)0.7
& 92.3\(\pm\)0.4
& 4.47\(\pm\)0.03
& 4.50\(\pm\)0.03 \\

InfoAgent
& \textbf{28.8\(\pm\)1.3}
& \textbf{96.4\(\pm\)0.3}
& \textbf{4.63\(\pm\)0.02}
& \textbf{4.62\(\pm\)0.02} \\
\bottomrule
\end{tabular}
}
\label{tab:supp_generation_robustness}
\end{table}

The small variation in ReqText-F1, Bind-Q, and Q-Align indicates that the main conclusions are stable across independent generations. Full varies more because it is a strict image-level binary endpoint, but InfoAgent remains strongest in every run.

\subsection{Planner Transfer}
\label{sec:supp_backbone}

We replace Gemini 3.1 Pro with Claude Opus 4.6 while preserving the complete 200-request test set, evidence bundles, design-prior bank, raster renderer, routing rules, verification protocol, and evaluation suite.

\begin{table}[t]
\centering
\caption{Planner transfer on the complete test set.}
\resizebox{\linewidth}{!}{
\begin{tabular}{@{}llrrrrr@{}}
\toprule
Planner & Executor & ReqCov & Full & ReqText-F1 & Bind-Q & Q-Align \\
\midrule
Gemini 3.1 Pro & Same-IVD & 81.00 & 21.5 & 86.4 & 4.58 & 4.522 \\
Gemini 3.1 Pro & InfoAgent & 83.54 & 28.5 & 96.5 & 4.64 & 4.621 \\
Claude Opus 4.6 & Same-IVD & 80.4 & 20.0 & 85.7 & 4.55 & 4.49 \\
Claude Opus 4.6 & InfoAgent & 82.8 & 27.0 & 95.8 & 4.61 & 4.60 \\
\bottomrule
\end{tabular}
}
\label{tab:supp_planner_transfer}
\end{table}

Executable IVD improves Full and required-text fidelity under both proprietary planners. We interpret this as planner transfer within strong closed-source systems, not as open-model or backbone-independent generalization.

\subsection{Evaluator Robustness and Human Audit}
\label{sec:supp_evaluator_audit}

The primary evaluator is compared with a Claude Opus 4.6 evaluator and a blinded human audit. Three annotators independently evaluate 80 complete outputs for Full, required factual items, and text--visual bindings; disagreements are resolved by majority vote. Output order and method identity are hidden.

\begin{table}[t]
\centering
\caption{Agreement with blinded human evaluation.}
\begin{tabular}{@{}lrrr@{}}
\toprule
Evaluator & Full \(\kappa\) & Fact F1 & Binding F1 \\
\midrule
Gemini 3.1 Pro & 0.76 & 91.4 & 87.2 \\
Claude Opus 4.6 & 0.73 & 90.8 & 86.5 \\
Human--human & 0.79 & 93.0 & 89.1 \\
\bottomrule
\end{tabular}
\label{tab:supp_evaluator_agreement}
\end{table}

Against the strongest competing output available for each request, human judges prefer InfoAgent in 50 of 80 cases, report 14 ties, and prefer the baseline in 16 cases. Cross-model disagreements concentrate on visually ambiguous bindings and paraphrased source-supported claims rather than exact symbolic content.

\section{IVD, Verification, and Repair Details}
\label{sec:supp_method_details}

\subsection{Executable IVD and Routing}
\label{sec:supp_ivd}

An IVD is \(z=(\mathcal{E},\mathcal{U},\mathcal{R},G,\Omega)\), where \(\mathcal{E}\) is the evidence pool, \(\mathcal{U}\) the regional canvas specification, \(\mathcal{R}\) the reading-flow graph, \(G=(\mathcal{A},\mathcal{D})\) the typed element and dependency graph, and \(\Omega\) the executable obligations. Each element stores metadata, payload and provenance, region and route, local bindings, and applied obligations.

Routing selects the smallest non-empty subset of \(\{\mathrm{visual},\mathrm{symbolic},\mathrm{binding}\}\) whose capabilities satisfy the declared obligations. The implementation enumerates the seven subsets. Visual execution handles open-ended backgrounds, metaphors, and illustrations; symbolic execution handles exact text, values, formulas, charts, legends, and sources; binding execution handles arrows, callouts, labels, and legend mappings. Symbolic and binding objects preserve rendering-tree handles, bounding boxes, z-order, payloads, evidence identifiers, and anchor geometry. Raster elements retain region traces; uncertain object-level anchors remain \textsc{Unknown}.

\paragraph{Design-prior construction and retrieval.}
The bank is constructed from topic-suppressed layout and style descriptions: named entities, factual values, source text, and topic-specific claims are removed before embedding and clustering. Online retrieval is content-aware. A query may mention the communication domain, information structure, and desired visual motifs when these cues determine an appropriate prior, but it cannot copy factual payloads, evidence spans, exact values, dates, or source-specific language. Retrieved cards therefore contribute reusable layout and aesthetic constraints rather than factual content.

\subsection{Selective Three-Valued Verification}
\label{sec:supp_verification}

For each obligation \(\omega_j\), checker \(\nu_j\) returns \((s_j,\kappa_j,w_j)\), where \(s_j\in\{\textsc{Pass},\textsc{Fail},\textsc{Unknown}\}\), \(\kappa_j\) is confidence, and \(w_j\) is a witness. Rendering-tree checkers inspect exact strings, values, formulas, clipping, overlap, contrast, z-order, and endpoint geometry. Evidence checkers compare payloads with cited spans, while visual-semantic checkers inspect only the traced object or crop.

Precision, recall, and F1 are computed over covered decisions, excluding \textsc{Unknown}; the abstention rate is reported separately. For certification, CritPass, and repair stopping, every critical \textsc{Unknown} is treated as non-pass and is never certified. It enters the repair queue only when the execution trace exposes a typed and bounded patch target; otherwise, it remains unresolved.

\begin{table}[t]
\centering
\caption{Checker audit on 600 obligation decisions.}
\resizebox{0.86\linewidth}{!}{
\begin{tabular}{@{}lrrrrr@{}}
\toprule
Checker & \(N\) & Prec. & Rec. & F1 & Unknown \\
\midrule
Rendering-tree
& 240 & 97.7 & 95.0 & 96.3 & 0.0 \\
Evidence support
& 150 & 92.5 & 84.1 & 88.1 & 5.3 \\
Binding geometry
& 110 & 90.8 & 79.8 & 84.9 & 2.7 \\
Visual semantic
& 100 & 89.9 & 76.5 & 82.7 & 10.0 \\
\bottomrule
\end{tabular}
}
\label{tab:supp_checker_audit}
\end{table}

Rendering-tree checks are the most reliable because they operate on editable objects and exact geometry. Model-assisted checks are less accurate and abstain more often, particularly for visual-semantic relations involving raster objects.

\paragraph{Coverage--risk trade-off.}
We vary the confidence thresholds used by model-assisted checkers. Coverage is the fraction of decisions receiving \textsc{Pass} or \textsc{Fail}, and selective risk is their error rate. Critical Unknown is the abstention rate over critical obligations; false repair is the percentage of committed repairs triggered by an incorrect verifier decision.

\begin{table}[t]
\centering
\caption{Coverage--risk trade-off for model-assisted verification.}
\resizebox{0.86\linewidth}{!}{
\begin{tabular}{@{}lrrrr@{}}
\toprule
Threshold
& Coverage
& Risk
& \shortstack{Critical\\Unknown}
& \shortstack{False\\repair} \\
\midrule
Relaxed
& 97.0 & 14.5 & 3.0 & 5.8 \\
Default
& 94.2 & 9.8 & 5.8 & 3.3 \\
Conservative
& 85.0 & 6.2 & 15.0 & 1.7 \\
\bottomrule
\end{tabular}
}
\label{tab:supp_coverage_risk}
\end{table}

The default threshold balances coverage and selective risk: domain-specific risk ranges from 8.1\% to 11.7\%. Over all 200 test outputs, the certificate produces 106 true passes, 9 false passes, 17 false non-passes, and 68 true non-passes, corresponding to 92.2\% precision and 86.2\% recall against external CritPass labels. Human agreement on the obligation-level audit is Fleiss' \(\kappa=0.82\).

\subsection{Dependency-Aware Scoped Repair}
\label{sec:supp_repair}

A violation \(q\) identifies an element, field, witness, severity, and initial patch class. Its dependency closure \(H(q)\) contains the elements that may require modification or re-verification. Let \(P(H)\) denote all previously passed obligations whose scopes do not intersect \(H(q)\). This protected set is fixed by the original dependency closure, even when a candidate patch later expands to a region or the full canvas.

Let \(\operatorname{NP}_{\mathrm{crit}}(\mathcal{C})=F^{\mathrm{crit}}+U^{\mathrm{crit}}\) be the number of critical non-pass obligations. A candidate certificate \(\mathcal{C}'\) is admissible when: (i) the targeted obligation becomes \textsc{Pass}, or \(\operatorname{NP}_{\mathrm{crit}}(\mathcal{C}')<\operatorname{NP}_{\mathrm{crit}}(\mathcal{C})\); (ii) every previously passed critical obligation remains \textsc{Pass}; and (iii) every obligation in \(P(H)\) remains \textsc{Pass}. The second condition prevents critical regressions within the affected closure, while the third protects both critical and non-critical content outside it.

Candidate operations include evidence-backed payload replacement, symbolic re-rendering, constrained layout reflow, anchor reassignment, masked visual editing, regional regeneration, and global recompilation. InfoAgent searches scopes in increasing order from field to element, dependency closure, region, and global recompilation. Within the first scope containing an admissible candidate, it minimizes \(Q^\star\), followed by the number of changed elements, edited canvas area, and tool cost.

Let \(\operatorname{Adm}(\mathcal{C}_p;\mathcal{C},q,P)\) denote the repair acceptance rule: the target becomes \textsc{Pass} or the number of critical non-pass obligations strictly decreases; every previously passed critical obligation remains \textsc{Pass}; and every obligation in \(P\) remains \textsc{Pass}. Among admissible candidates at the first successful scope, \(\operatorname{Best}\) minimizes \(Q^\star\), followed by changed elements, edited area, and tool cost.

\begin{algorithm}[t]
\small
\caption{Dependency-aware scoped repair}
\label{alg:supp_repair}
\KwIn{IVD \(z\), output \(y\), trace \(T\), certificate \(\mathcal{C}\), rounds \(K\)}
\KwOut{Updated \(y,T,\mathcal{C}\)}
\For{\(k\leftarrow1\) \KwTo \(K\)}{
    \(\mathsf{progress}\leftarrow\mathrm{false}\)\;
    \(\mathcal{V}\leftarrow\) current critical non-pass obligations, ordered by severity\;
    \If{\(\mathcal{V}=\varnothing\)}{\textbf{break}}

    \ForEach{\(q\in\mathcal{V}\)}{
        \If{\(s_q(\mathcal{C})=\textsc{Pass}\)}{\textbf{continue}}
        \(H\leftarrow H(q)\), \(P\leftarrow P(H)\)\;

        \ForEach{\(S\in\{\mathrm{field},\mathrm{element},\mathrm{closure},
        \mathrm{region},\mathrm{global}\}\)}{
            \(\mathcal{A}\leftarrow\varnothing\)\;
            \ForEach{\(p\in\operatorname{Generate}(q,S)\)}{
                \((y_p,T_p,\mathcal{C}_p)\leftarrow
                \operatorname{ExecuteVerify}(p,y,T,\mathcal{C},q,P)\)\;
                \eIf{\(\operatorname{Adm}(\mathcal{C}_p;\mathcal{C},q,P)\)}{
                    \(\mathcal{A}\leftarrow\mathcal{A}\cup
                    \{(p,y_p,T_p,\mathcal{C}_p)\}\)\;
                }{
                    rollback \(p\)\;
                }
            }
            \If{\(\mathcal{A}\neq\varnothing\)}{
                \((p^\star,y,T,\mathcal{C})\leftarrow
                \operatorname{Best}(\mathcal{A})\)\;
                commit \(p^\star\); \(\mathsf{progress}\leftarrow\mathrm{true}\)\;
                \textbf{break}\tcp*[r]{leave scope loop}
            }
        }
    }
    \If{\(\mathsf{progress}=\mathrm{false}\)}{
        \textbf{break}\tcp*[r]{no admissible patch this round}
    }
}
\Return{\(y,T,\mathcal{C}\)}
\end{algorithm}

The online verifier detects 90.6\% of externally confirmed critical violations. Localized repair fixes 85.9\% of correctly detected violations, yielding 77.8\% end-to-end repair recall. Four of 120 repaired outputs contain a newly introduced violation (NewImg \(=3.3\%\)); seven regressions among 875 previously passed obligations outside the original dependency closure give CollatReg \(=0.8\%\). The mean union of committed edits is 12.4\% of the canvas.

\section{Additional Qualitative Results}
\label{sec:supp_qualitative}

\begin{figure*}[t]
\centering
\includegraphics[width=\textwidth]{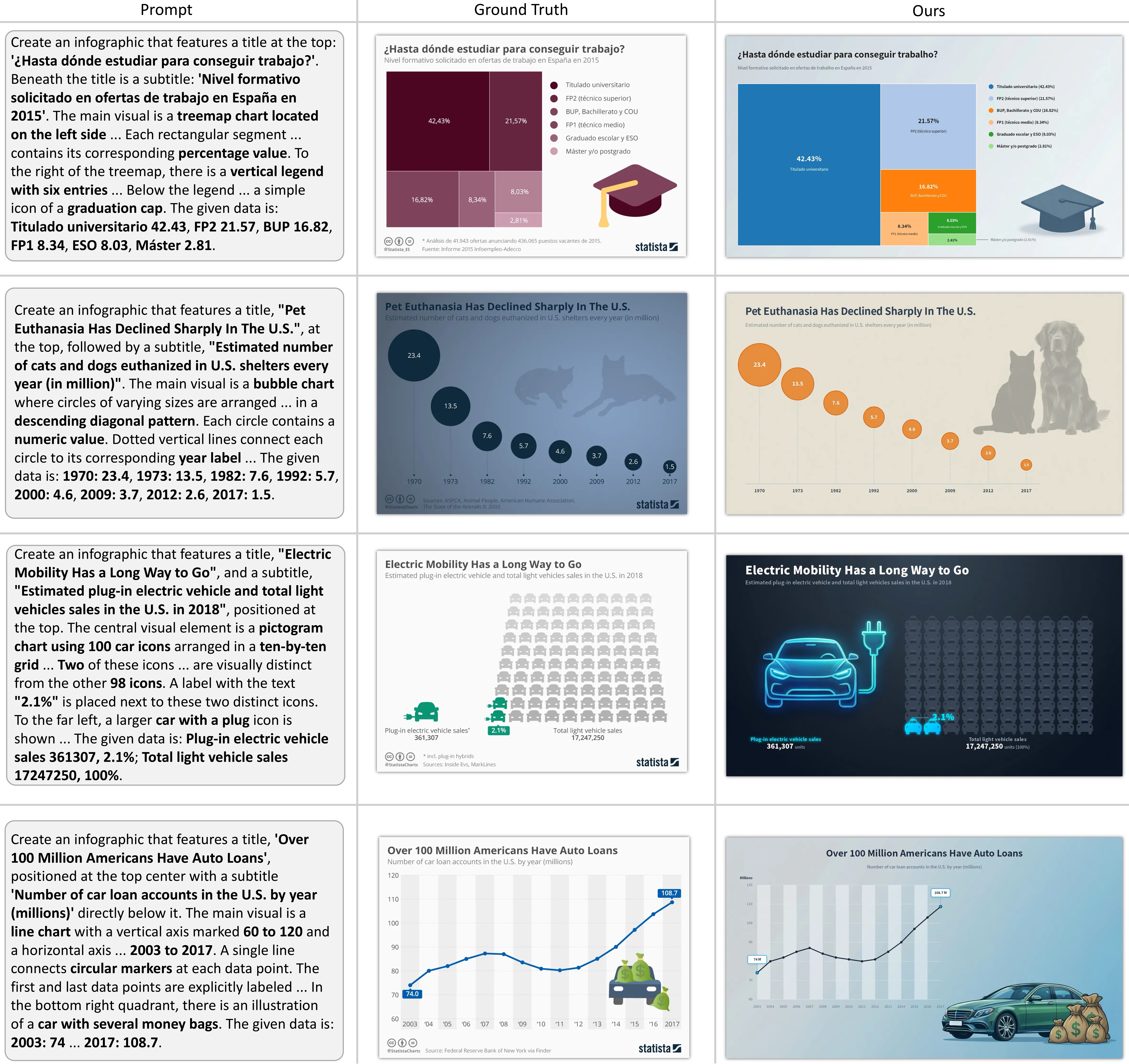}
\caption{\textbf{Additional IGenBench results.}
Each row presents the benchmark prompt, reference infographic, and InfoAgent output. The examples cover treemap, bubble-chart, pictogram, and line-chart layouts and are shown without manual correction.}
\label{fig:supp_igenbench_examples}
\end{figure*}

\begin{figure*}[t]
\centering
\includegraphics[width=\textwidth]{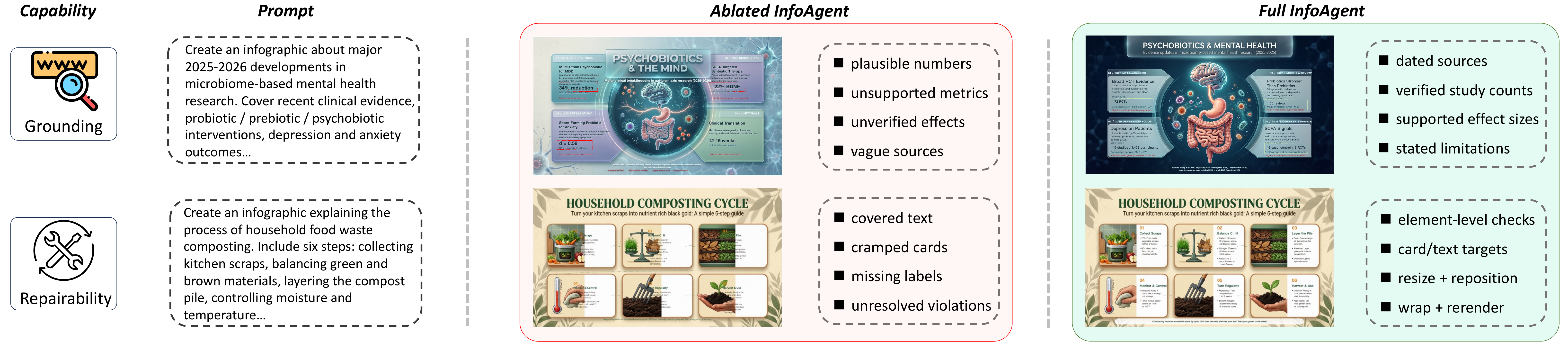}
\caption{\textbf{Additional evidence-controlled comparisons.}
The examples compare factual coverage, symbolic fidelity, local grounding, and visual hierarchy across generation methods, and include representative unresolved and unsuccessful-repair cases.}
\label{fig:supp_additional_comparisons}
\end{figure*}

This section supplements the aggregate results with two qualitative views. Figure~\ref{fig:supp_igenbench_examples} evaluates structural fidelity across diverse IGenBench formats under the prompt-only protocol. Figure~\ref{fig:supp_additional_comparisons} examines evidence-controlled generation across methods, focusing on factual coverage, symbolic precision, local text--visual grounding, and visual organization. The selected examples also include unresolved and unsuccessful-repair cases to illustrate the remaining limitations of the system.

The IGenBench examples show that InfoAgent can realize markedly different visual structures while preserving the requested chart organization and symbolic content. In the evidence-controlled comparisons, direct raster generation often produces plausible global compositions but remains less stable on exact values and dense labels. Search augmentation improves contextual coverage without consistently preserving element-level bindings, whereas code-oriented rendering strengthens symbolic structure at the cost of more rigid compositions. InfoAgent combines evidence-bound payloads with visual, symbolic, and binding execution. Its remaining failures primarily involve uncertain raster anchors, incomplete or conflicting evidence, and layout defects that require broader replanning rather than a local patch.

\clearpage

\par
\switchonecolumn

\section{Evaluation Prompts}
\label{sec:supp_evaluation_prompts}

\begin{tcolorbox}[
    colback=lightgray!10,
    colframe=black,
    title={\textbf{Checklist and Five-Dimension Evaluation Prompts}},
    breakable
]

\begin{Verbatim}[breaklines=true,breakanywhere=true,breaksymbolleft={},fontsize=\tiny]
A. EVIDENCE-GROUNDED CHECKLIST EVALUATION

System Prompt

You are a strict and impartial evaluator of evidence-grounded
infographics.

You will receive:

1. The original generation request.
2. A generated infographic.
3. A frozen evidence bundle.
4. A held-out checklist of atomic requirements.

The checklist and evidence bundle are evaluation-only annotations.
They are not available to the generator, IVD compiler, online verifier,
or repair procedure.

Your task is to evaluate each checklist item independently.

Evaluation principles:

- Judge only content visibly present in the generated infographic.
- Use only the supplied evidence bundle for factual verification.
- Do not use external knowledge or infer the author's intention.
- Do not inspect or assume access to IVD fields, execution traces,
  checker witnesses, repair logs, or intermediate outputs.
- A visually absent required item is FAIL.
- A visible claim contradicted by the evidence is FAIL.
- A visible claim without sufficient evidence support is FAIL.
- If an item cannot be judged because the relevant text, relation,
  or object is unreadable or ambiguous, return UNKNOWN.
- Do not assume that unreadable text is correct.
- Do not penalize stylistic differences unless they violate an
  explicit checklist requirement.
- Exact required-string matching is computed separately by the
  required-text evaluator. Do not replace that metric with semantic
  guessing.
- Report additional critical factual, symbolic, binding, or layout
  errors that are visibly present but not explicitly listed in the
  checklist.

Status definitions:

PASS:
The visible output satisfies the requirement and, where applicable,
is supported by the supplied evidence.

FAIL:
The requirement is missing, contradicted, unsupported, visibly
incorrect, or bound to the wrong object.

UNKNOWN:
The available visual or evidential information is insufficient to
determine PASS or FAIL.

Output only one valid JSON object:

{
  "items": [
    {
      "item_id": "C1",
      "status": "PASS | FAIL | UNKNOWN",
      "visible_witness": "Concise description of the visible evidence",
      "evidence_id": "E1 or null",
      "reason": "Brief justification"
    }
  ],
  "additional_critical_errors": [
    {
      "type": "factual | symbolic | binding | layout",
      "description": "Visible error not already represented by an item"
    }
  ],
  "critical_pass": true,
  "all_pass": true
}

critical_pass is true only when every item marked critical is PASS and
no additional critical error is present.

all_pass is true only when every checklist item is PASS and no
additional critical error is present.

Do not output Markdown or text outside the JSON object.


Example Checklist Instance: bench_00001

Topic:
A public-health infographic explaining the gut-brain axis.

Atomic requirements:

C1 [critical, content]
The infographic presents the gut-brain axis as bidirectional
communication among the gastrointestinal system, enteric nervous
system, gut microbiota, and central nervous system.

C2 [critical, content]
The infographic identifies four major signaling pathways:
neural, immune, endocrine/HPA-axis, and metabolic.

C3 [critical, binding]
The vagus nerve is visually and textually associated with neural
signaling between the gut and the brain.

C4 [critical, factual]
The infographic states that gut microorganisms can produce or
modulate serotonin (5-HT), GABA, and dopamine.

C5 [critical, factual]
The infographic states that approximately 90% of the body's serotonin
is synthesized in the gut by enterochromaffin cells.

C6 [critical, qualification]
The 90% statement is not presented as meaning that 90% of brain
serotonin originates in the gut, because peripheral serotonin does
not directly cross the blood-brain barrier.

C7 [non-critical, factual]
Representative GABA-producing taxa include Lactobacillus,
Bifidobacterium, and Bacteroides, with the GAD pathway noted when
space permits.

C8 [critical, content]
Dysbiosis is associated with HPA-axis dysregulation, inflammation,
increased intestinal permeability, neuroinflammation, depression,
or anxiety.

C9 [critical, content]
Short-chain fatty acids are presented as mediators of immune and
neural signaling.

C10 [non-critical, factual]
Psychobiotic effects are described as strain-specific, with examples
such as L. rhamnosus, B. longum, or L. gasseri.

C11 [non-critical, content]
Diet, antibiotic exposure, and psychological stress are presented as
factors that can influence the gut-brain axis.

C12 [non-critical, advice]
A general adult fiber recommendation may be shown as an evidence-based
nutrition suggestion, but it must not be presented as a gut-brain-axis-
specific clinical threshold.


B. GEMINI 3.1 PRO FIVE-DIMENSION QUALITY EVALUATION

System Prompt

You are a strict and impartial expert in infographic design and
visual communication.

You will receive:

1. The original generation prompt.
2. One generated infographic.

Evaluate the following five dimensions independently:

1. Prompt Compliance
2. Text Readability
3. Information Organization and Layout
4. Text-Visual Consistency and Binding
5. Visual Design Quality

General principles:

- Evaluate only content visibly present in the image.
- Do not infer the generation process or the author's intention.
- Score each dimension independently.
- For Prompt Compliance, evaluate only requirements explicitly stated
  in the original prompt.
- Do not add requirements that are absent from the prompt.
- Domain-knowledge coverage and factual support are evaluated by the
  separate evidence-grounded checklist and must not be re-evaluated
  here.
- Exact required-string, number, date, and citation matching are
  evaluated separately. Here, judge only whether text is visibly
  readable.
- Additional content should remain neutral unless it is off-topic,
  misleading, contradictory to the prompt, or materially harms
  readability.
- Do not use external knowledge for open-world factual checking.
- Every score must be supported by visible evidence in the image.
- Do not compute an overall score.

Dimension definitions:

1. Prompt Compliance

Evaluate whether the output follows the explicitly requested topic,
language, audience, infographic format, canvas organization, structure,
and visual style.

Do not duplicate domain-knowledge completeness evaluation.

2. Text Readability

Evaluate whether titles, body text, labels, values, and source notes
are legible. Consider font size, contrast, line spacing, density,
garbled characters, overlap, clipping, and boundary violations.

Do not perform exact string matching in this dimension.

3. Information Organization and Layout

Evaluate hierarchy, sectioning, reading order, alignment, spacing,
negative space, density, and overall page balance.

4. Text-Visual Consistency and Binding

Evaluate whether text and visual elements support one another and
whether labels, arrows, legends, colors, symbols, and callouts are
clearly associated with the intended objects.

Judge only relations visible inside the infographic.

5. Visual Design Quality

Evaluate color harmony, typography, illustrations, icons, background,
stylistic consistency, professionalism, visual appeal, and overall
finish.

Do not duplicate layout or knowledge-completeness judgments.

Scoring scale:

1: Severe failure; largely unusable.
2: Multiple major visible problems.
3: Usable, but with clear room for improvement.
4: Strong overall quality with only minor local issues.
5: Excellent quality with almost no visible problem.

Use scores 1 and 5 conservatively.

User Prompt Template

Original generation prompt:

{generation_prompt}

Generated infographic:

{generated_infographic}

Evaluate the five dimensions and output only one valid JSON object:

{
  "prompt_compliance": {
    "score": 1,
    "evidence": ["Concrete visible observation"],
    "reason": "Brief explanation"
  },
  "text_readability": {
    "score": 1,
    "evidence": ["Concrete visible observation"],
    "reason": "Brief explanation"
  },
  "information_organization_and_layout": {
    "score": 1,
    "evidence": ["Concrete visible observation"],
    "reason": "Brief explanation"
  },
  "text_visual_consistency_and_binding": {
    "score": 1,
    "evidence": ["Concrete visible observation"],
    "reason": "Brief explanation"
  },
  "visual_design_quality": {
    "score": 1,
    "evidence": ["Concrete visible observation"],
    "reason": "Brief explanation"
  }
}

Every score must be an integer from 1 to 5.

If no concrete evidence is available for a dimension, return an empty
evidence array.

Do not output an overall score, Markdown, or any text outside the JSON.
\end{Verbatim}
\end{tcolorbox}
\section{Main Pipeline Prompt and Case Trace}
\label{sec:supp_main_prompt}

\begin{tcolorbox}[
    colback=lightgray!10,
    colframe=black,
    title={\textbf{InfoAgent Prompt, Condensed Trace, and Representative IVD Fragment}},
    breakable
]

\begin{Verbatim}[breaklines=true, breakanywhere=true, breaksymbol={}, fontsize=\tiny, samepage=false]
A. SHARED PIPELINE SPECIFICATION

InfoAgent contains three execution roles:

1. IVD Planner and Compiler
2. Visual-Layer Generator
3. Symbolic and Binding Renderer

A separate scoped verifier evaluates the composed result and returns
element-indexed violations for repair.

The IVD is an executable multimodal specification rather than a longer
generation prompt. It does not contain low-level SVG code, but each
information-bearing element records its payload, provenance, region,
execution route, local dependencies, bindings, and verification
obligations.

Canvas dimensions and orientation are specified by each IVD rather
than fixed globally. The representative case below uses:

width: 1080 px
height: 1350 px
orientation: portrait


B. IVD PLANNER AND COMPILER SYSTEM PROMPT

You are the IVD Planner and Compiler in an infographic-generation
pipeline.

Transform a user request, retrieved evidence, and one retrieved design
prior into an executable Infographic Visual Description (IVD).

Do not generate images, SVG, HTML, or CSS.

Workflow:

1. Determine whether external factual evidence is required.
2. When required, call text_search to retrieve reliable evidence.
3. Call get_skill exactly once.
4. Compile the request, evidence spans, and retrieved design prior into
   one executable IVD.
5. Output the IVD and stop.

Tool budget:

- text_search: optional, at most three calls
- get_skill: required, exactly one call
- total tool calls: at most four

Design-prior retrieval:

The get_skill query must be written in English and emphasize visual
flow, section structure, panel arrangement, color tone, atmosphere,
and graphic style.

The query may mention the communication domain, content category, and
desired visual motifs when they help identify an appropriate design
prior. For example, it may request a film-editorial dashboard or
letter-related visual motifs.

The query must not contain:

- named entities from the request;
- copied factual claims;
- exact values, dates, ratings, or rankings;
- evidence spans or source-specific wording;
- text that must appear verbatim in the final infographic.

The retrieved skill contributes layout and aesthetic constraints only;
it must not supply factual content.

Factual rules:

- Do not invent facts, numbers, dates, rankings, prices, proportions,
  sources, study results, or quotations.
- Exact factual strings must be copied from user-provided material or
  retrieved evidence.
- Paraphrased claims must retain supporting evidence identifiers.
- When sources disagree, prefer authoritative evidence and qualify
  claims when necessary.
- If reliable evidence is unavailable, use a qualitative statement or
  omit the claim.

The IVD must contain:

canvas:
- width, height, and orientation

evidence_pool:
- evidence identifiers and source spans used by elements

regions:
- region identifiers, approximate locations, owners, purposes, and
  flexible ranges

reading_flow:
- ordered regions or typed precedence relations

elements:
- element id
- type and semantic role
- priority
- exact or paraphrasable payload
- evidence provenance
- assigned region
- primary execution route
- optional auxiliary routes
- local bindings
- executable obligations

dependencies:
- contains
- precedes
- supports
- binds_to
- depends_on

Allowed execution routes:

- visual
- symbolic
- binding
- any necessary combination of these routes

Routing principles:

- Open-ended backgrounds, metaphors, textures, illustrations, and
  non-critical visual objects use the visual route.
- Exact text, values, dates, formulas, charts, legends, and source
  notes use the symbolic route.
- Arrows, leader lines, callouts, label-object relations, and legend
  mappings use the binding route.
- Hybrid routes are used only when one route cannot satisfy all
  declared obligations.

Executable obligations include:

- exact
- support
- time_scope
- contrast
- minimum_font_size
- containment
- no_overlap
- endpoint_inside
- binding_consistency

For each tool call, provide only a one-sentence action rationale.
Do not expose private chain-of-thought.

Final output:

<answer>
<ivd>{valid JSON}</ivd>
</answer>

Do not output additional text.


C. VISUAL-LAYER GENERATOR ADDENDUM

You are the Visual-Layer Generator.

Input:

- executable IVD;
- elements whose routes contain visual;
- visual regions and clean symbolic regions;
- retrieved style prior;
- optional visual references.

Generate only:

- background;
- main visual subject;
- complex illustrations;
- visual metaphors;
- textures;
- non-critical icons;
- display elements explicitly assigned to visual execution.

Do not generate:

- small body text;
- exact numerical values;
- detailed formulas;
- source notes;
- dense labels;
- fake charts;
- unrequested logos or watermarks.

Keep symbolic and binding regions visually quiet and sufficiently clean
for subsequent editable rendering.

Return:

- image_path;
- asset_id;
- region-level execution traces;
- confidence for model-assisted object grounding.

At most two visual refinements are allowed, and only when a required
visual element is missing, the style is inconsistent, or a symbolic
region is visually polluted.


D. SYMBOLIC AND BINDING RENDERER ADDENDUM

You are the Symbolic and Binding Renderer.

Input:

- executable IVD;
- generated visual layer;
- elements routed to symbolic or binding execution;
- canvas dimensions and regional constraints.

Render symbolic content and binding geometry as editable SVG objects.

Exact strings must be copied directly from protected IVD payload fields.
The renderer may determine typography, wrapping, placement, and local
geometry, but must not rewrite factual payloads.

Symbolic objects include:

- titles assigned to symbolic execution;
- body text;
- values and units;
- dates;
- formulas;
- charts;
- legends;
- labels;
- source notes.

Binding objects include:

- arrows;
- leader lines;
- callouts;
- label-object anchors;
- legend-object mappings.

Every rendered object must retain:

- IVD element id;
- protected payload;
- bounding box;
- z-order;
- assigned region;
- anchor geometry when applicable;
- evidence identifier;
- rendering-tree handle.

Binding geometry is deterministic once its source and target anchors
are resolved. If a raster target cannot be grounded confidently, return
UNKNOWN rather than guessing.

The final SVG must reference the visual-layer image and must not contain
diagnostic grids or implementation annotations.


E. SCOPED VERIFICATION AND REPAIR ADDENDUM

For every obligation, return:

- PASS, FAIL, or UNKNOWN;
- confidence;
- witness;
- checker version;
- element id;
- affected field;
- patch target.

Rendering-tree checks inspect exact strings, values, formulas,
containment, clipping, overlap, contrast, z-order, and endpoint geometry.

Evidence checks compare factual payloads with their cited evidence
spans.

Visual-semantic checks inspect only the traced object or crop.

A failed or critical-UNKNOWN obligation becomes an element-indexed
violation.

Repair begins with the smallest valid scope:

field -> element -> dependency closure -> region -> global recompilation

A candidate patch is accepted only when:

- the targeted critical obligation becomes PASS, or the total number
  of critical non-pass obligations decreases;
- no unrelated critical PASS obligation becomes non-pass;
- previously passed obligations outside the target dependency closure
  remain passed.

Otherwise, the candidate is rolled back and the scope is expanded.


F. CONDENSED LOGGED CASE TRACE

The following trace is condensed from one execution log. Tool inputs
and returned factual fields are preserved. Action rationales are
paraphrased for readability and do not expose hidden model reasoning.

User Request

Generate an infographic explaining the release, cast, box-office
trajectory, and critical reception of the film "Dear You".


Action 1: Evidence Retrieval

Rationale:

Release information, names, box-office values, audience counts, and
reception claims require external evidence before IVD compilation.

<tool_call>
{
  "name": "text_search",
  "arguments": {
    "queries": [
      "Dear You film release date runtime director cast",
      "Dear You May 2026 box office trajectory",
      "Dear You box office milestone audience count",
      "Dear You critical reception rating"
    ]
  }
}
</tool_call>

<tool_response>
E1:
Release date: April 30, 2026.
Runtime: 118 minutes.

E2:
Director: Lan Hongchun.
Principal cast: Li Sitong, Wang Yantong, Wu Shaoqing,
Zheng Runqi, and Wang Xiaohui.

E3:
Comparison box-office values for other May releases:
486 million yuan and 293 million yuan.

E4:
The film exceeded one billion yuan and 31 million viewers
by May 24, 2026.

E5:
The cited reception snapshot reports an audience rating of 9.1/10.
...
</tool_response>


Action 2: Content-Aware Design-Prior Retrieval

Rationale:

The request benefits from a data-rich editorial layout combining
metric cards, a trajectory chart, cast information, and film-and-letter
visual motifs.

<tool_call>
{
  "name": "get_skill",
  "arguments": {
    "query": "data-dashboard mosaic for a film box-office infographic
              with a title band, metric cards, a central trajectory
              chart, cast and reception side panels, a source strip,
              an editorial navy and cream palette, muted red and gold
              accents, and restrained film-and-letter motifs"
  }
}
</tool_call>

<tool_response>
Layout skill:
data-dashboard-mosaic with a title band, metric cards, central chart
region, side information panels, and bottom source strip.

Style skill:
clean editorial blueprint using navy, cream, muted red, and gold,
with restrained film, letter, stamp, and floral motifs.
</tool_response>


Action 3: IVD Compilation

Rationale:

The retrieved evidence and design prior specify the factual payloads,
spatial regions, routes, bindings, and obligations required by the
request.

Representative IVD Fragment

The full logged IVD contains all requested elements. The fragment below
shows representative elements for release, cast, box-office trajectory,
audience count, reception, and source attribution. Decorative assets,
minor chart labels, auxiliary layout nodes, and repeated style fields
are omitted for space. This display fragment is not passed separately
to the executor.

<ivd_fragment>
{
  "canvas": {
    "width": 1080,
    "height": 1350,
    "orientation": "portrait"
  },
  "evidence_pool": {
    "E1": "release date and runtime",
    "E2": "director and principal cast",
    "E3": "comparison box-office values",
    "E4": "box-office milestone and audience count",
    "E5": "reception snapshot"
  },
  "regions": [
    {
      "id": "title_band",
      "owner": "symbolic",
      "position": "top full-width"
    },
    {
      "id": "hero_visual",
      "owner": "visual",
      "position": "upper-left"
    },
    {
      "id": "release_cast_panel",
      "owner": "symbolic",
      "position": "upper-right"
    },
    {
      "id": "trajectory_panel",
      "owner": "symbolic",
      "position": "middle full-width"
    },
    {
      "id": "reception_panel",
      "owner": "symbolic",
      "position": "lower-right"
    },
    {
      "id": "source_strip",
      "owner": "symbolic",
      "position": "bottom full-width"
    }
  ],
  "reading_flow": [
    "title_band",
    "hero_visual",
    "release_cast_panel",
    "trajectory_panel",
    "reception_panel",
    "source_strip"
  ],
  "elements": [
    {
      "id": "value_release_date",
      "type": "date",
      "role": "release_metric",
      "priority": "critical",
      "payload": {
        "text": "April 30, 2026",
        "mode": "exact",
        "evidence": "E1"
      },
      "region": "release_cast_panel",
      "route": {
        "primary": "symbolic",
        "auxiliary": []
      },
      "bindings": [],
      "obligations": [
        "exact",
        "support",
        "time_scope",
        "contrast",
        "containment",
        "no_overlap"
      ]
    },
    {
      "id": "cast_block",
      "type": "body_text",
      "role": "cast_information",
      "priority": "critical",
      "payload": {
        "text": "Director: Lan Hongchun. Cast: Li Sitong,
                 Wang Yantong, Wu Shaoqing, Zheng Runqi,
                 and Wang Xiaohui.",
        "mode": "exact",
        "evidence": "E2"
      },
      "region": "release_cast_panel",
      "route": {
        "primary": "symbolic",
        "auxiliary": []
      },
      "bindings": [],
      "obligations": [
        "exact",
        "support",
        "minimum_font_size",
        "containment",
        "no_overlap"
      ]
    },
    {
      "id": "value_boxoffice_milestone",
      "type": "value",
      "role": "key_metric",
      "priority": "critical",
      "payload": {
        "text": "Over 1 billion yuan",
        "mode": "exact",
        "evidence": "E4"
      },
      "region": "trajectory_panel",
      "route": {
        "primary": "symbolic",
        "auxiliary": ["binding"]
      },
      "bindings": [
        {
          "type": "binds_to",
          "target": "boxoffice_trajectory"
        }
      ],
      "obligations": [
        "exact",
        "support",
        "contrast",
        "containment",
        "no_overlap",
        "endpoint_inside"
      ]
    },
    {
      "id": "value_audience_count",
      "type": "value",
      "role": "audience_metric",
      "priority": "critical",
      "payload": {
        "text": "31 million viewers",
        "mode": "exact",
        "evidence": "E4"
      },
      "region": "trajectory_panel",
      "route": {
        "primary": "symbolic",
        "auxiliary": []
      },
      "bindings": [],
      "obligations": [
        "exact",
        "support",
        "containment",
        "no_overlap"
      ]
    },
    {
      "id": "boxoffice_trajectory",
      "type": "chart",
      "role": "trajectory",
      "priority": "critical",
      "payload": {
        "evidence": ["E3", "E4"]
      },
      "region": "trajectory_panel",
      "route": {
        "primary": "symbolic",
        "auxiliary": ["binding"]
      },
      "bindings": [],
      "obligations": [
        "support",
        "label_consistency",
        "containment",
        "no_overlap"
      ]
    },
    {
      "id": "reception_rating",
      "type": "value",
      "role": "reception_metric",
      "priority": "critical",
      "payload": {
        "text": "Audience rating: 9.1/10",
        "mode": "exact",
        "evidence": "E5"
      },
      "region": "reception_panel",
      "route": {
        "primary": "symbolic",
        "auxiliary": []
      },
      "bindings": [],
      "obligations": [
        "exact",
        "support",
        "time_scope",
        "contrast",
        "containment"
      ]
    },
    {
      "id": "source_note",
      "type": "source_note",
      "role": "attribution",
      "priority": "critical",
      "payload": {
        "text": "Sources: 1905.com; People's Daily;
                 Southern Daily; cited rating snapshot.",
        "mode": "exact",
        "evidence": ["E1", "E2", "E3", "E4", "E5"]
      },
      "region": "source_strip",
      "route": {
        "primary": "symbolic",
        "auxiliary": []
      },
      "bindings": [],
      "obligations": [
        "exact",
        "support",
        "minimum_font_size",
        "containment",
        "no_overlap"
      ]
    }
  ],
  "dependencies": [
    {
      "source": "value_boxoffice_milestone",
      "relation": "binds_to",
      "target": "boxoffice_trajectory"
    },
    {
      "source": "value_audience_count",
      "relation": "supports",
      "target": "boxoffice_trajectory"
    },
    {
      "source": "source_note",
      "relation": "supports",
      "target": "release_cast_panel"
    },
    {
      "source": "source_note",
      "relation": "supports",
      "target": "trajectory_panel"
    },
    {
      "source": "source_note",
      "relation": "supports",
      "target": "reception_panel"
    }
  ]
}
</ivd_fragment>


Action 4: Visual-Layer Execution

Rationale:

The visual layer requires film-poster, letter, postage, and floral
references while preserving clean regions for exact symbolic content.

<tool_call>
{
  "name": "image_search",
  "arguments": {
    "queries": [
      "Dear You film poster",
      "vintage overseas Chinese letter envelope",
      "Chinese postage stamp and postmark texture",
      "red kapok blossom branch"
    ]
  }
}
</tool_call>

<tool_response>
IMG_001: film-poster reference
IMG_002: letter and postage reference
IMG_003: floral motif reference
...
</tool_response>

<tool_call>
{
  "name": "image_generation",
  "arguments": {
    "reference_images": ["IMG_001", "IMG_002", "IMG_003"],
    "prompt": "Create a portrait editorial film-infographic visual
               layer with vintage letter texture, muted red blossoms,
               navy and cream tones, a strong poster motif in the
               upper-left, and quiet clean regions for release, cast,
               trajectory, reception, and source information.
               Include no small factual text or fake labels."
  }
}
</tool_call>

<tool_response>
image_path: IMAGE_LAYER_001.png
visual_trace:
  hero_visual -> upper-left region
  decorative_letter -> background region
  floral_motif -> upper and lower margins
</tool_response>


Action 5: Symbolic and Binding Execution

Rationale:

Protected release, cast, box-office, audience, reception, and source
payloads are rendered as editable objects, together with the trajectory
chart and its local bindings.

The full renderer output contains all IVD objects. The excerpt below
lists the handles corresponding to the displayed fragment.

<answer>
{
  "svg_path": "SYMBOLIC_BINDING_LAYER_001.svg",
  "trace": {
    "value_release_date": "svg:text#release_date",
    "cast_block": "svg:g#cast_block",
    "value_boxoffice_milestone": "svg:text#boxoffice_value",
    "value_audience_count": "svg:text#audience_count",
    "boxoffice_trajectory": "svg:g#trajectory_chart",
    "reception_rating": "svg:text#reception_rating",
    "source_note": "svg:text#source_note"
  }
}
</answer>


Action 6: Composition, Verification, and Scoped Repair

Compose:

IMAGE_LAYER_001.png
+
SYMBOLIC_BINDING_LAYER_001.svg
->
COMPOSITE_001.png

Verifier output:

{
  "element_id": "value_boxoffice_milestone",
  "obligation": "containment",
  "status": "FAIL",
  "confidence": 1.0,
  "witness": "The final word 'yuan' extends 14 px beyond the
              metric-card boundary.",
  "patch_target": "symbolic",
  "scope": "element"
}

Repair action:

Preserve the exact payload "Over 1 billion yuan".
Reduce the font size from 34 px to 31 px and reflow the same text
within the existing metric-card region.

Re-check:

- exact payload equality;
- containment;
- minimum font size;
- contrast;
- overlap;
- the binding to boxoffice_trajectory;
- protected obligations outside the dependency closure.

Final status:

{
  "element_id": "value_boxoffice_milestone",
  "status": "PASS",
  "payload_changed": false,
  "font_size_before": 34,
  "font_size_after": 31,
  "edited_scope": "single symbolic element",
  "unrelated_critical_regressions": 0
}
\end{Verbatim}
\end{tcolorbox}
\section{Design-Prior Skill Synthesis}
\label{sec:supp_skill_prompts}

\begin{tcolorbox}[
    colback=lightgray!10,
    colframe=black,
    title={\textbf{Layout and Style Skill Synthesis Prompts}},
    breakable
]
\vspace{0.5em}
\begin{Verbatim}[breaklines=true, breaksymbol={}, fontsize=\tiny]
A. LAYOUT SKILL SYNTHESIS

System Prompt

You are a senior infographic design-systems expert with expertise in
visual composition, spatial hierarchy, reading flow, and information
architecture.

Your task is to synthesize a reusable layout skill from a cluster of
topic-suppressed infographic descriptions.

The skill must capture spatial regularities rather than topic-specific
content. It should be sufficiently concrete for a downstream planner
to allocate regions, assets, and reading order on a new topic.

User Prompt Template

Below are {n} infographic descriptions that share a common spatial
layout pattern.

The first {top_n} descriptions are the examples closest to the cluster
centroid. The final {boundary_n} descriptions are boundary examples
that represent the variation range of the cluster.

--- TYPICAL EXAMPLES ---

{typical_block}

--- BOUNDARY EXAMPLES ---

{boundary_block}

Synthesize exactly one layout skill file in the following format:

# {kebab-case-name}

{One-line description of the spatial pattern.}

## Visual Flow

{Describe how the eye moves through the layout and where hierarchy and
visual tension originate. Use one to three sentences.}

## Structure

- {Primary spatial division of the canvas}
- {Dominant regions and relative visual weights}
- {Use of negative space and separation}

## Asset Zones

- {zone-name}: {position and approximate proportion};
  {number and type of assets}; {primary, secondary, or optional weight}
- {Repeat for each distinct zone}

## Best For

{Infographic structures and content organizations for which the layout
is most suitable. Use one sentence.}

Requirements:

- The name must contain two to four kebab-case words.
- The name must describe the spatial pattern, not the topic.
- Asset Zones must use concrete spatial terms such as top, bottom,
  left, right, center, full-width, or approximate proportions.
- Describe reusable layout behavior rather than the content of the
  examples.
- Do not introduce facts or text copied from a specific example.
- Output only the skill file in Markdown.
- Do not output explanations or code fences.


Representative Layout Skill

# anchored-axis-panel-spread

A dominant directional axis anchors the composition while supporting
information panels extend into the surrounding margins.

## Visual Flow

The eye enters through a strong title block and moves toward a central
horizontal or vertical spine. Sequential nodes establish the primary
reading path, while peripheral panels provide contextual detail and
visual rest points.

## Structure

- A central axis occupies approximately 20-30% of the canvas and
  separates the title/context region from the detailed information
  region.
- The title and sequential axis carry the dominant visual weight;
  corner panels and callouts remain secondary.
- Negative space is limited but preserved between axis nodes and
  panel boundaries.

## Asset Zones

- title-block: upper-left to upper-center; one title and one subtitle;
  primary weight
- context-panel: upper-right; one introductory card or quotation;
  secondary weight
- central-axis: middle full-width band; four to seven sequential nodes;
  primary weight
- supplementary-left: lower-left; one or two detail panels;
  secondary weight
- supplementary-right: lower-right; one or two summary panels;
  secondary weight
- decorative-anchors: corners or axis endpoints; one to three thematic
  illustrations; optional weight

## Best For

Timelines, process explanations, and evolution narratives that require
a strong directional spine with contextual panels around it.


B. STYLE SKILL SYNTHESIS

System Prompt

You are a senior visual-design expert specializing in infographic
aesthetics, color theory, typography, illustration, and visual branding.

Your task is to synthesize a reusable style skill from a cluster of
topic-suppressed infographic descriptions.

The skill must capture aesthetic regularities independently of layout
and content topic.

User Prompt Template

Below are {n} infographic descriptions that share a common visual
style.

The first {top_n} descriptions are the examples closest to the cluster
centroid. The final {boundary_n} descriptions represent the stylistic
variation within the cluster.

--- TYPICAL EXAMPLES ---

{typical_block}

--- BOUNDARY EXAMPLES ---

{boundary_block}

Synthesize exactly one style skill file in the following format:

# {kebab-case-name}

{One-line description of the visual mood.}

## Color Tone

{Describe temperature, saturation, contrast, and dominant hue families
in one or two sentences. Do not use hexadecimal values.}

## Visual Atmosphere

{Describe the emotional and professional impression produced by the
style in one or two sentences.}

## Key Visual Elements

- {Most distinctive concrete visual element}
- {Second concrete visual element}
- {Third concrete visual element}
- {Optional fourth element}
- {Optional fifth element}

## Best For

{Themes, domains, and communication tones for which the style is most
suitable. Use one sentence.}

Requirements:

- The name must contain two to four kebab-case words.
- The name must describe the visual style, not the content domain.
- Color Tone must be expressed in natural-language color terms.
- Key Visual Elements must name concrete components rather than
  abstract adjectives.
- Do not infer a spatial layout unless it is itself a stylistic motif.
- Output only the skill file in Markdown.
- Do not output explanations or code fences.


Representative Style Skill

# flat-modular-infographic

Clean, structured knowledge presentation using icon-paired modules and
soft muted color families.

## Color Tone

A neutral-warm cream or light-beige foundation is combined with
desaturated blue, sage green, muted yellow, and dusty pink accents.
Titles use deep navy or charcoal to establish moderate contrast.

## Visual Atmosphere

The style resembles a carefully designed handbook or educational
reference page. It feels calm, trustworthy, approachable, and suitable
for dense information without excessive visual fatigue.

## Key Visual Elements

- Modular color-coded cards arranged as grids or sequential flows
- Small line-style or flat-fill icons paired with concise labels
- Strong title hierarchy with restrained secondary typography
- Numbered steps or arrow-connected process nodes
- Generous internal spacing between compact information groups

## Best For

Educational explainers, practical guides, process diagrams, and
knowledge-dense reference infographics requiring a calm professional
appearance.
\end{Verbatim}
\end{tcolorbox}

\par
\switchtwocolumn


\end{document}